\documentclass[pdflatex,sn-mathphys-num]{sn-jnl}

\usepackage{graphicx}%
\usepackage{multirow}%
\usepackage{amsmath,amssymb,amsfonts}%
\usepackage{amsthm}%
\usepackage{mathrsfs}%
\usepackage[title]{appendix}%
\usepackage{xcolor}%
\usepackage{textcomp}%
\usepackage{manyfoot}%
\usepackage{booktabs}%
\usepackage{algorithm}%
\usepackage{algorithmicx}%
\usepackage{algpseudocode}%
\usepackage{listings}%
\usepackage{rotating}
\usepackage{multirow}
\usepackage{booktabs}
\usepackage[figuresright]{rotating}
\usepackage{makecell}
\usepackage{svg}

\renewcommand{\figurename}{Figure}

\theoremstyle{thmstyleone}%
\theoremstyle{thmstyletwo}%

\theoremstyle{thmstylethree}%

\begin{document}

\title[Article Title]{A Multimodal Foundation Model to Enhance Generalizability and Data Efficiency for Pan-cancer Prognosis Prediction}


\author[1]{\fnm{Huajun} \sur{Zhou}}\email{csehjzhou@ust.hk}

\author[1]{\fnm{Fengtao} \sur{Zhou}}\email{fzhouaf@connect.ust.hk}

\author[1]{\fnm{Jiabo} \sur{Ma}}\email{jmabq@cse.ust.hk}

\author[1]{\fnm{Yingxue} \sur{Xu}}\email{yxueb@connect.ust.hk}

\author[1]{\fnm{Xi} \sur{Wang}}\email{vancywangxi@ust.hk}

\author[2]{\fnm{Xiuming} \sur{Zhang}}\email{1508056@zju.edu.cn}

\author[3,4,5]{\fnm{Li} \sur{Liang}}\email{lli@smu.edu.cn}

\author[6]{\fnm{Zhenhui} \sur{Li}}\email{lizhenhui@kmmu.edu.cn}

\author*[1,7,8,9,10]{\fnm{Hao} \sur{Chen}}\email{jhc@cse.ust.hk}

\affil[1]{\orgdiv{Department of Computer Science and Engineering}, \orgname{Hong Kong University of Science and Technology}, \orgaddress{\city{Hong Kong}, \country{China}}}
\affil[2]{Department of Pathology, The First Affiliated Hospital, School of Medicine, Zhejiang University, Hangzhou, China}%
\affil[3]{Department of Pathology, Nanfang Hospital and School of Basic Medical Sciences, Southern Medical University, Guangzhou, China}
\affil[4]{Guangdong Provincial Key Laboratory of Molecular Tumor Pathology, Guangzhou, China}
\affil[5]{Jinfeng Laboratory, Chongqing, China}
\affil[6]{Department of Radiology, The Third Affiliated Hospital of Kunming Medical University, Yunnan Cancer Hospital, Kunming, China}%
\affil[7]{Department of Chemical and Biological Engineering, Hong Kong University of Science and Technology, Hong Kong, China}%
\affil[8]{Division of Life Science, Hong Kong University of Science and Technology, Hong Kong, China}%
\affil[9]{HKUST Shenzhen-Hong Kong Collaborative Innovation Research Institute, Futian, Shenzhen, China}%
\affil[10]{State Key Laboratory of Nervous System Disorders, Hong Kong University of Science and Technology, Hong Kong, China}%




\abstract{
Multimodal data provides heterogeneous information for a holistic understanding of the tumor microenvironment.
However, existing AI models often struggle to harness the rich information within multimodal data and extract poorly generalizable representations.
Here we present MICE (Multimodal data Integration via Collaborative Experts), a multimodal foundation model that effectively integrates pathology images, clinical reports, and genomics data for precise pan-cancer prognosis prediction. 
Instead of conventional multi-expert modules, MICE employs multiple functionally diverse experts to comprehensively capture both cross-cancer and cancer-specific insights. 
Leveraging data from 11,799 patients across 30 cancer types, we enhanced MICE’s generalizability by coupling contrastive and supervised learning.
MICE outperformed both unimodal and state-of-the-art multi-expert-based multimodal models, demonstrating substantial improvements in C-index ranging from 3.8\% to 11.2\% on internal cohorts and 5.8\% to 8.8\% on independent cohorts, respectively. 
Moreover, it exhibited remarkable data efficiency across diverse clinical scenarios.
With its enhanced generalizability and data efficiency, MICE establishes an effective and scalable foundation for pan-cancer prognosis prediction, holding strong potential to personalize tailored therapies and improve treatment outcomes.
}


\maketitle

\section{Introduction}\label{sec1}
Cancer remains a major global health burden, accounting for one in six deaths worldwide \cite{bray2024global}. 
In precision oncology, accurate prognosis prediction is essential for guiding clinical decisions, including therapy selection, care pathway optimization, and resource allocation. 
For individual patients, it enables risk-adapted triage, helps tailor treatment intensity, and facilitates timely access to targeted supportive or palliative care.
Conversely, inaccurate predictions may risk delaying critical interventions that preserve quality of life for cancer patients \cite{sparano2018adjuvant, hui2021importance, orlovic2023accuracy}.
To improve prognostic accuracy, the primary challenge lies in the intrinsic biological complexity of tumor heterogeneity, manifesting at both molecular and histological levels. 
Overcoming this challenge necessitates a comprehensive characterization of tumor microenvironment (TME), a vital determinant of tumor evolution, therapeutic vulnerabilities, and treatment resistance \cite{boehm2022multimodal, bortolini2021multimodal, gao2024explainable}.

Artificial intelligence (AI)-powered multimodal learning holds transformative potential for comprehensively characterizing the TME \cite{zhou2024multimodal, nakach2024comprehensive, boehm2022harnessing}.
Substantial recent studies \cite{zhang2024prototypical, byeon2025interpretable, song2024multimodal, volinsky2024prediction, xiang2024development, xiong2024mome, qian2024multimodal, sharma2025hybrid, bai2025predicting, xu2024multimodal} have introduced advanced multimodal AI models that integrate histopathology images and genomics data to improve prognostic accuracy.
Despite their impressive performance, these models are typically trained on cancer-type-specific multimodal datasets, inherently limiting sample size and compromising model's generalizability.
In contrast, multimodal foundation models (MFMs) can learn representative features from large-scale datasets, thereby substantially enhancing their generalizability.
Existing MFMs \cite{lu2024visual, xiang2025vision, huang2023visual} primarily leverage contrastive learning to align features across modalities from the same patient and often function only as unimodal feature extractors in downstream clinical applications.  
Yet, owing to substantial inter-modal heterogeneity \cite{wang2023shared, shao2023fam3l}, these MFMs tend to capture limited common knowledge while overlooking valuable modality-specific information.
Therefore, an MFM that can provide an effective and generalized multimodal data integration by comprehensively capturing both shared and unique information across modalities for clinical translation remains an unmet need.

Developing an MFM necessitates a large-scale multimodal dataset for pre-training, which is a major obstacle in precision oncology.
Although multimodal data are scarce for individual cancer types, different cancers share underlying patterns of correlation across modalities such as pathology images, genomics, and clinical reports.
This consistent inter-modal relevance suggests that aggregating multimodal data across cancer types represents a viable and compelling strategy for building a unified MFM capable of precise and generalizable prognosis prediction.

\begin{figure*}[!t]
\centering
    \includegraphics[width=0.99 \textwidth]{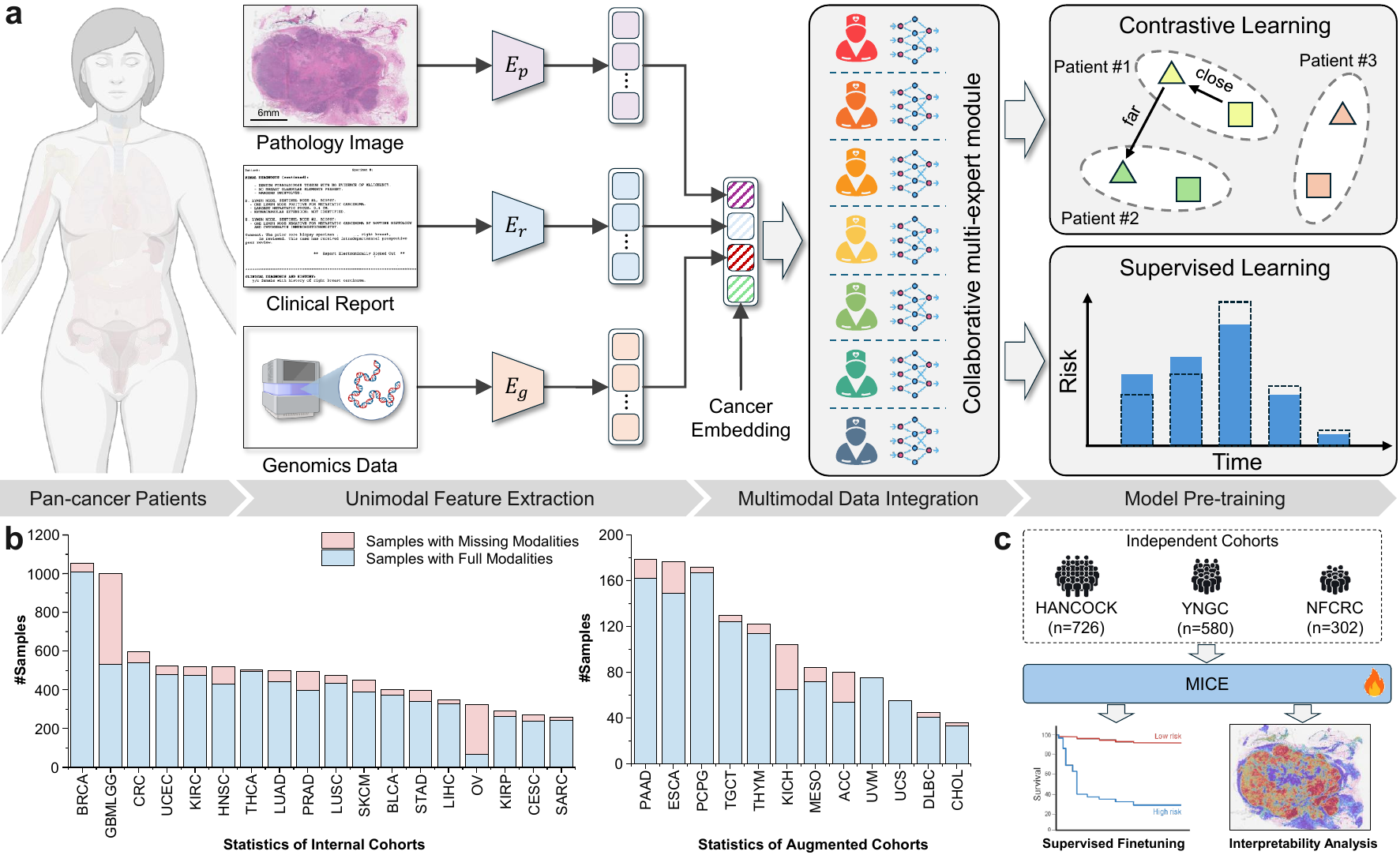}
\caption{\textbf{Development and validation of the multimodal foundation model MICE for effective multimodal data integration.}
\textbf{a}, The overall pipeline of MICE.
We curated multimodal data, including whole-slide images, clinical reports, and molecular profiles, from 11,799 patients spanning 30 cancer types for model development and validation.
MICE employs a tailored network architecture comprising modality-specific encoders for unimodal feature extraction and a collaborative multi-expert module for effective multimodal data integration.
MICE is pre-trained by leveraging the dual contrastive and supervised learning strategy, jointly optimizing feature alignment and prognosis relevance to capture both generalizable and discriminative patterns within multimodal data.
\textbf{b}, TCGA cohort partition for model development.
Multimodal data from 10,191 patients across 30 TCGA cancer types were categorized into 1) 18 internal cohorts (each $n\geq200$) for pre-training and cross-validation and 2) 12 augmented cohorts (each $n\textless200$) for pre-training only, ensuring robust representation learning and mitigating small-sample bias.
\textbf{c}, Clinical translation via supervised finetuning.
The pre-trained MICE was adapted to three independent cohorts (n=726, 580, and 302), enabling precise prognosis prediction across diverse patient populations.
}
\label{fig:framework}
\end{figure*}

Effectively leveraging pan-cancer data to develop MFMs requires an advanced architecture capable of comprehensively capturing both shared and cancer-specific biological patterns.
Critically, while different cancers exhibit common prognostic features, each also contains cancer-specific biological insights.
Conventional multi-head models \cite{shao2024multi} employ multiple expert modules to capture specialized knowledge per cancer type but often overlook valuable inter-cancer relationships that could enhance feature discrimination.
In contrast, Mixture-of-Experts (MoE) models \cite{xiong2024mome, zhou2022mixture, jiang2024m4oe} leverage cross-cancer correlations through a routing module to dynamically select experts based on patients’ multimodal data, yet frequently capture insufficient cancer-specific knowledge.
Consequently, current approaches focus exclusively on either shared or cancer-specific patterns, lacking the integrated representation essential for accurate pan-cancer prognosis.


In this study, we present MICE (Multimodal data Integration via Collaborative Experts), to our knowledge, the first MFM explicitly designed to effective integrate heterogeneous whole slide images (WSIs), clinical reports, and genomics data for accurate pan-cancer prognosis prediction.
As illustrated in Figure \ref{fig:framework}a, MICE was developed leveraging multimodal data from 11,799 patients across 30 cancer types, curated from two public datasets and three collaborating hospitals.
MICE incorporates a collaborative multi-expert module that captures inter-cancer correlations while preserving cancer-specific biological insights through three distinct expert groups: an overlapping MoE-based group for cross-cancer patterns via input-conditioned routing, a specialized group to extract cancer-specific knowledge, and a consensual expert to integrate shared patterns across all cancers. 
Together, these experts collaboratively extract a holistic representation essential for generalizable pan-cancer prognosis.
Furthermore, we pre-trained MICE using a dual learning strategy combining contrastive and supervised learning on large-scale pan-cancer datasets to enhance generalizability. 
MICE was extensively validated on 18 internal (n=8,932) and 10 independent (n=1,608) prognosis prediction tasks. 
MICE significantly outperformed both unimodal and state-of-the-art multi-expert multimodal models, with C-index improvements of 3.8\% to 11.2\% on internal cohorts and 5.8\% to 8.8\% on independent cohorts, demonstrating excellent generalizability. 
It also demonstrated strong data efficiency, achieving performance comparable to existing models trained on full datasets even when fine-tuned with 50\% fewer samples.
With enhanced generalizability and data efficiency, MICE establishes an effective and scalable foundation for precise pan-cancer prognosis prediction, with the potential to support personalized therapy decisions and improve treatment outcomes.

\section{Results}
\subsection{Study cohort}
We curated multimodal data comprising 11,799 pan-cancer patients sourced from public TCGA and HANCOCK \cite{doerrich2024multimodal} datasets, alongside three collaborating hospitals in China.
This includes 12,876 WSIs, 11,535 clinical reports, and 9,486 genomic profiles in total.
Detailed cohort statistics are provided in Supplementary Tables \ref{tab:internal} and \ref{tab:external}.

\textbf{TCGA dataset for model development.} 
The TCGA dataset includes 11,254 WSIs, 9,927 clinical reports, 9,486 genomic profiles from 10,191 patients across 30 cohorts, while each cohort corresponding to a specific cancer type (Figure \ref{fig:framework}b).
Since several TCGA cohorts have too few patients to yield reliable internal cross-validation results, and therefore we split 30 TCGA cohorts as follows:
1) 18 internal validation cohorts (n=8,932), each containing more than 200 patients, were used for pre-training and internal validation following the five-fold cross-validation protocol. 
By default, we employed overall survival (OS) records as supervised signals. 
For three cohorts (TCGA-KIRP, TCGA-PRAD, TCGA-THCA), we used progression-free survival (PFS) because they contained fewer than 50 patients with uncensored OS records, which could compromise statistical reliability.
2) 12 augmented cohorts (n=1,259), each comprising fewer than 200 patients, were used exclusively for the model’s pre-training stage and excluded from validation because of their limited sample sizes.

\textbf{Independent cohorts.}
We collected three out-of-domain cohorts (n=1,608) for independent validation (Figure \ref{fig:framework}c).
1) HANCOCK \cite{doerrich2024multimodal} (HANC, n=726): Public head and neck cancer cohort with paired WSIs, surgery reports, and five follow-up records, including OS, PFS, disease-free survival (DFS), disease-specific survival (DSS), and distant metastasis-free survival (DMFS);
2) YNGC (n=580): A retrospective collection of an in-house gastric cancer cohort comprising 580 cases from two hospitals, namely The First Affiliated Hospital and The Third Affiliated Hospital of Kunming Medical University. 
This cohort includes paired WSIs, pathology reports, and three follow-up records: OS, DMFS, and local recurrence-free survival (LRFS);
3) NFCRC (n=302): A retrospective colorectal cancer cohort retrospectively collected from Nanfang Hospital with paired WSIs, pathology reports, and two outcomes (OS, DSS).
We pre-train MICE using all samples in the TCGA dataset, and finetune it on independent cohorts following the five-fold cross-validation protocol.

\begin{figure*}[!t]
\centering
    \includegraphics[width=0.99 \textwidth]{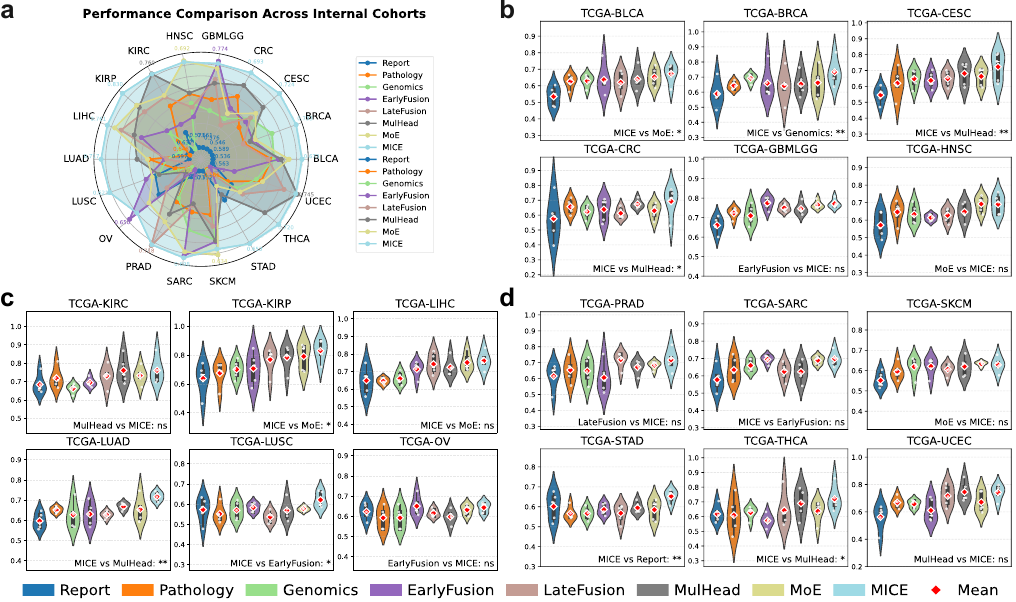}
\caption{\textbf{Model performance of prognosis prediction on internal cohorts.} 
\textbf{a}, An overview of prognosis prediction performance across 18 internal cohorts. 
\textbf{b-d}, C-index scores on each internal cohort. White circles indicate the C-index scores of different folds. We perform the significance test between the best and second-best model on bottom right of each subfigure. ns: not significant; *: $p<0.05$; **: $p<0.01$.}
\label{fig:pretrain}
\end{figure*}

\begin{figure*}[!t]
\centering
    \includegraphics[width=0.99 \textwidth]{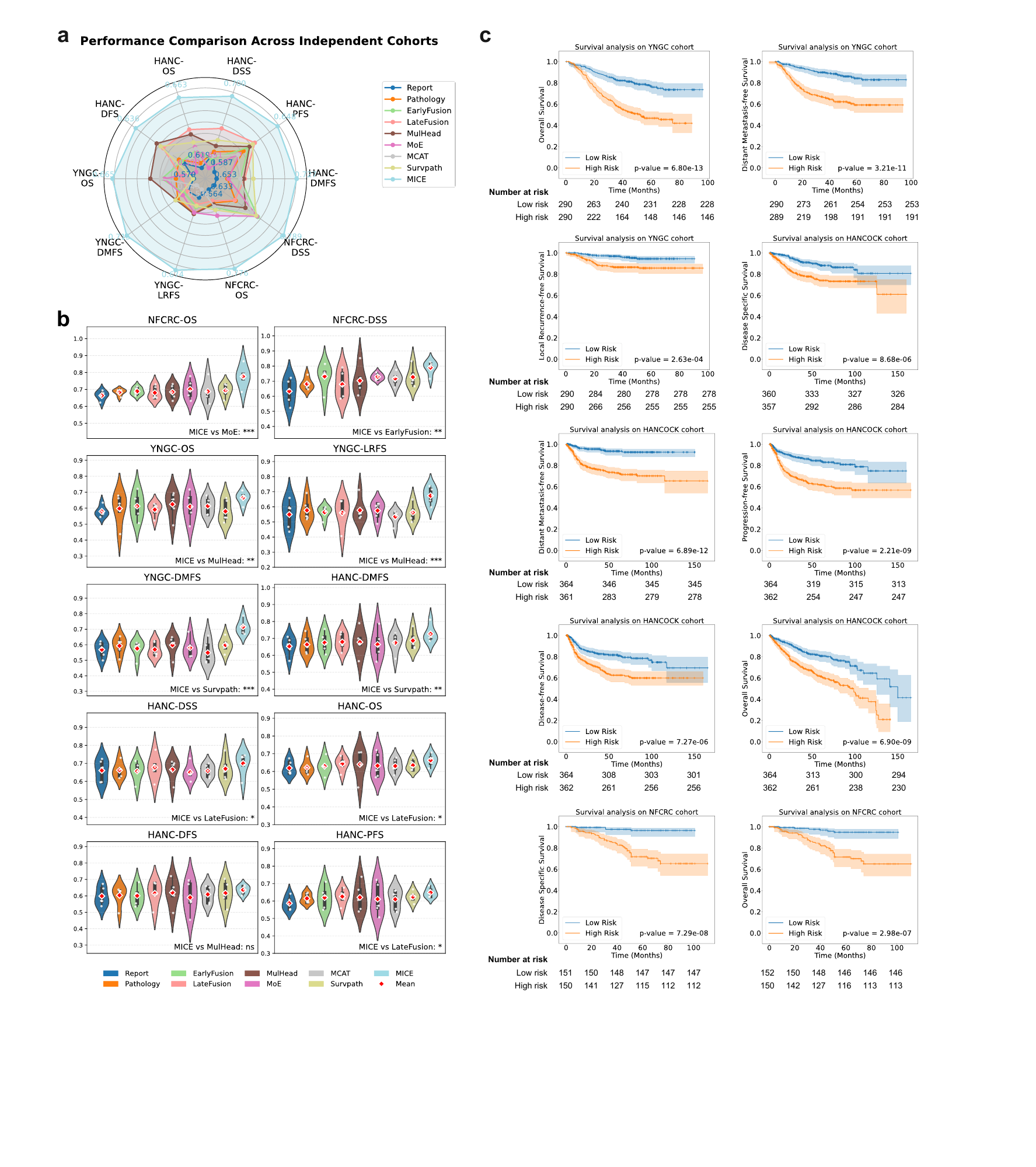}
\caption{\textbf{Model performance of prognosis prediction on independent cohorts.} 
\textbf{a}, An overview of performance across 10 independent prognosis prediction tasks.
\textbf{b}, C-index scores on each prognosis prediction task. White circles indicate the C-index scores of different folds. We perform the significance test between the best and second-best model on bottom right of each subfigure. ns: not significant; *: $p<0.05$; **: $p<0.01$; ***: $p<0.001$.
\textbf{c}, Kaplan–Meier curves of MICE on 10 prognosis prediction tasks.
}
\label{fig:external}
\end{figure*}

\subsection{MICE demonstrates strong prognostic accuracy on internal data}
To comprehensively evaluate MICE’s prognostic performance, we compared it against a diverse set of baseline methods, including three unimodal models (pathology-only, genomics-only, and report-only) and four representative multimodal models (early fusion, late fusion, multi-head \cite{shao2024multi}, and Mixture-of-Experts (MoE) \cite{jiang2024m4oe}).
Detailed results can be found in Table \ref{tab:res_internal}.
Figure \ref{fig:pretrain} showcases the remarkable performance of MICE, achieving an average C-index of 0.710 across 18 internal TCGA cohorts. 
Notably, our model demonstrated significant improvements over all competitors, surpassing multimodal models by 3.8\% to 6.4\% ($p<0.01$) and unimodal models by 7.0\% to 11.2\% ($p<0.001$) in terms of average C-index score.
The exceptional performance of MICE underscores its superiority over other multimodal AI models, highlighting the effectiveness of our collaborative multi-expert module and coupled contrastive and supervised pre-training. 
These two techniques enable the extraction of synergistic prognosis patterns from multimodal data, which are challenging to capture with existing multimodal AI models.
On an individual cohort level, MICE showcased remarkable generalizability across diverse cancer types, achieving the highest prognosis accuracy in 11 out of 18 cohorts (as depicted in Figure \ref{fig:pretrain}b-d). 
Among them, MICE exhibited a significant difference on 9 cohorts compared to the second-best models.
In addition, MICE achieved the second-best performance in the remaining 7 cohorts, demonstrating no statistically significant difference when compared to the best models in these cohorts.
Taken together, these results establish MICE as a new benchmark in multimodal pan-cancer prognosis prediction, showcasing its potential to advance precision oncology.

\subsection{MICE showcases excellent generalizability across independent cohorts}
Given that all independent cohorts contain only two modalities (i.e., WSIs and clinical reports), we further include two state-of-the-art bimodal models, MCAT \cite{mcat} and SurvPath \cite{jaume2024modeling}, for comparison. 
Detailed results are illustrated in Table \ref{tab:res_external}.
Across three independent cohorts, MICE consistently exhibits exceptional generalization capabilities. 
With an average C-index of 0.699 observed across 10 distinct prognosis prediction tasks within three independent cohorts (as depicted in Figure \ref{fig:external}a), MICE exhibited substantial absolute improvements ranging from 5.8\% to 8.8\% ($p < 0.001$) compared to its competitors.
Notably, MICE consistently outperformed MCAT and SurvPath, securing the top rank across all 10 prognosis prediction tasks, as illustrated in Figure \ref{fig:external}b. 
This consistency underscores MICE's robust predictive accuracy across diverse clinical scenarios and patient populations.
The clinical utility of MICE's risk stratification was further validated by Kaplan-Meier analysis in Figure \ref{fig:external}c.
Patients stratified into high-risk groups exhibited significantly shorter survival than low-risk counterparts (log-rank $p < 0.001$). 
This analysis underscores MICE's capacity to identify general survival patterns from multimodal data, generalizing effectively across cancer types, institutions, and incomplete modality inputs.

\begin{figure*}[!t]
\centering
    \includegraphics[width=0.99 \textwidth]{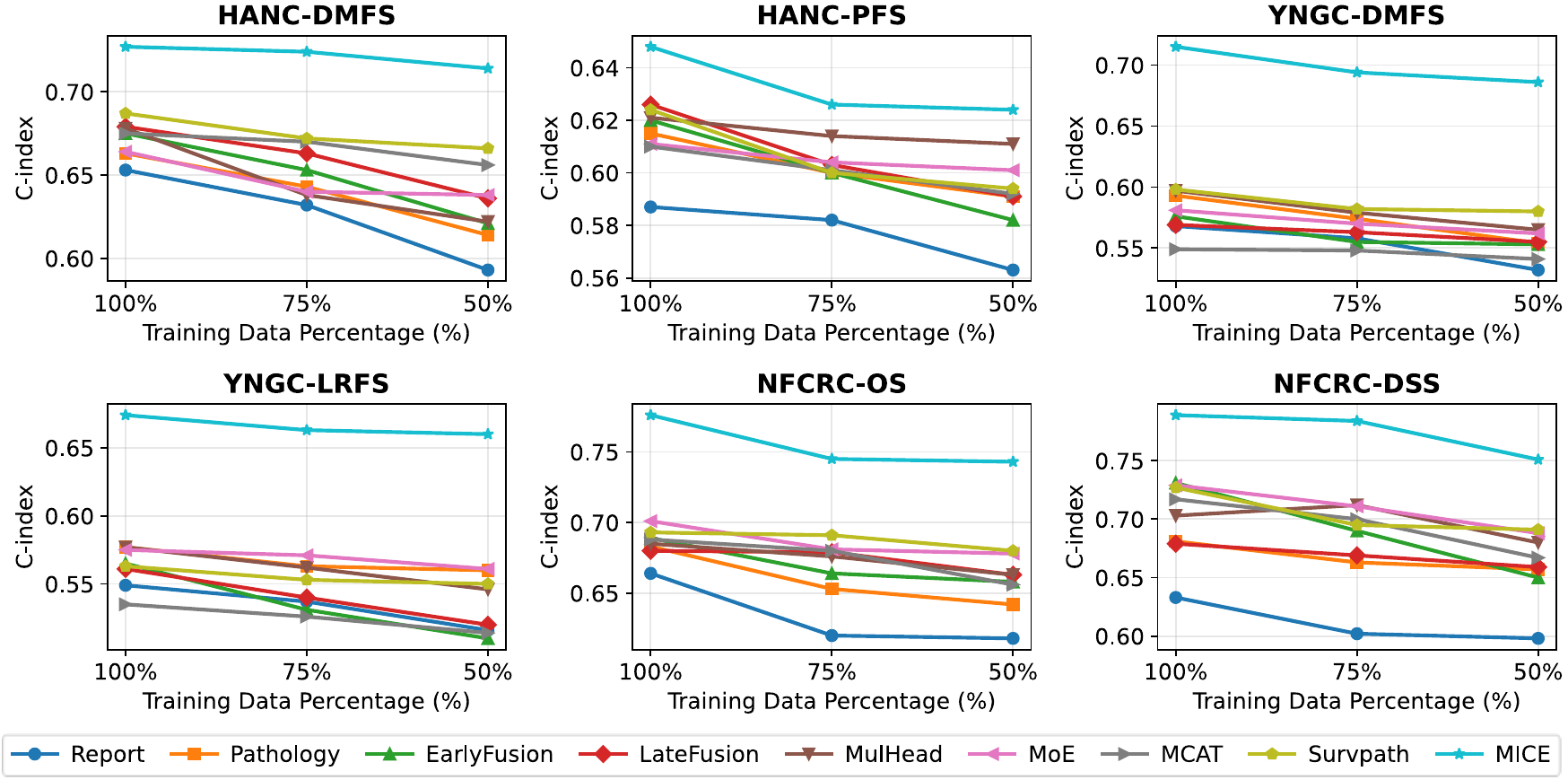}
\caption{\textbf{Model performance on data efficiency experiments.} We further train all compared models on independent prognosis prediction tasks using 50\% or 75\% of samples.}
\label{fig:label}
\end{figure*}

\subsection{MICE exhibits impressive data efficiency for real-world translation}
The formidable challenges of acquiring multimodal data from cancer patients underscore the pressing need for data-efficient AI models in clinical settings. 
To address this need, we conducted a comparative evaluation of MICE against unimodal and multimodal AI models by uniformly reducing the data scale for finetuning, specifically finetuning all models with 50\% or 75\% of the samples while keeping the test sets identical.
As shown in Figure \ref{fig:label}, MICE maintained similar performance margins to other models under both 50\% or 75\% of the available finetuning samples.
Most notably, MICE trained on just 50\% of the data achieved comparable performance against both unimodal and multimodal competitors trained on the full 100\% samples. 
MICE’s resilience to data scarcity underscores its potential to overcome the bottleneck in multimodal data collection.
These findings emphasize MICE's pivotal role in bridging the gap between AI innovation and clinical deployment, showcasing its potential to improve patient care by offering reliable performance even with limited multimodal data. 
This capability positions MICE as a crucial asset for translating cutting-edge AI technologies into practical clinical solutions, particularly in settings where comprehensive multimodal datasets are challenging to collect.

\subsection{Ablation study on different modality combinations}
\begin{figure*}[!t]
\centering
    \includegraphics[width=0.99 \textwidth]{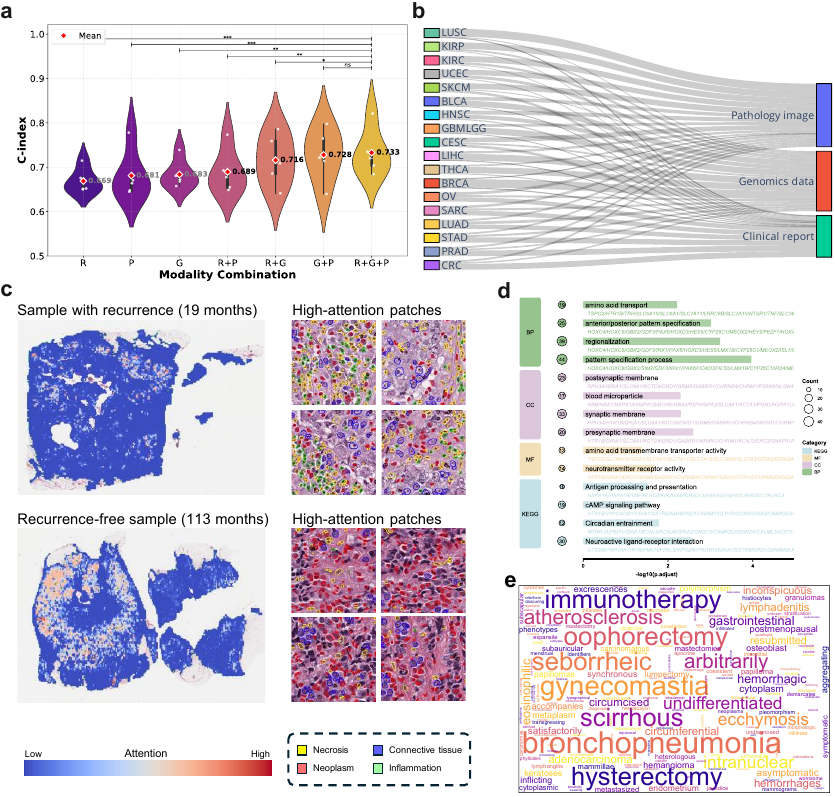}
\caption{\textbf{Interpretability analysis of MICE on the TCGA-BRCA cohort.} 
\textbf{a}, Performance of MICE model with different modalities as input. R: clinical report; P: pathology image; G: genomics data. The red points indicate the average C-index score of each model. ns: not significant; *: $p<0.05$; **: $p<0.01$; ***: $p<0.001$.
\textbf{b}, The contributions of each modality to the final prediction in each cohort.
\textbf{c}, Important WSI patches for prognosis prediction. Cell types are predicted by HoverNet \cite{graham2019hover}.
\textbf{d}, Gene ontology analysis on important genes. BP: biological processes; CC: cellular components; MF: molecular functions; KEGG: kyoto encyclopedia of genes and genomes.
\textbf{e}, Word cloud figure to visualize crucial words in clinical reports, where the font size is proportional to each word's predictive relevance.
}
\label{fig:visual}
\end{figure*}

To assess the individual contributions of each modality to multimodal data integration, we systematically investigated how the addition of pathology images (P), clinical reports (R), and genomics data (G) impacts the performance of prognosis prediction in MICE. 
Ablation studies were conducted across three tiers: 1) Unimodal: individual modalities (R, P, G); 2) Bimodal: pairwise combinations (R+P, R+G, P+G); 3) Trimodal: full integration of all three modalities (R+P+G). 
The evaluation of performance was carried out on the largest BRCA cohort in TCGA, with all three modalities available.
The results depicted in Figure \ref{fig:visual}a support several important conclusions. 
Firstly, prognosis accuracy exhibited a consistent increase with the inclusion of additional modalities, with the full integration achieving the highest C-index performance of 0.733±0.069. 
Secondly, all bimodal models outperformed their unimodal counterparts, with G+P (0.728±0.053) approaching performance scores comparable to the trimodal model. 
Thirdly, models incorporating genomics data as inputs consistently outperformed configurations incorporating other modalities, underscoring the paramount importance of genomics data in predicting patient prognosis within the TCGA-BRCA cohort.
Collectively, these findings quantitatively validate the integrative design of MICE, affirming its ability to extract synergistic prognosis patterns from multimodal data.

\subsection{Interpretability analysis reveals biologically and clinically salient prognosis drivers}
Interpretability is critical for AI-driven prognosis tools to gain trust from clinicians and patients. 
We validated MICE's decision-making through dual lenses: modality-level contribution analysis across cancer types and feature-level saliency mapping.

Figure \ref{fig:visual}b quantifies modality contributions across 18 cancer types. 
Pathology images demonstrated dominant contributions in cancers like kidney renal clear cell carcinoma (KIRC), kidney renal papillary cell carcinoma (KIRP), and lung squamous cell carcinoma (LUSC), reflecting histology's prognosis primacy. 
Genomic data shown stronger predictive value in molecularly-defined subtypes like glioblastoma/glioma (GBMLGG), liver hepatocellular carcinoma (LIHC), and breast invasive carcinoma (BRCA), corroborating our earlier finding in Figure \ref{fig:visual}a where genomic features are crucial for prognosis prediction in BRCA cohort. 
Clinical reports emerged as most influential in stomach adenocarcinoma (STAD), lung adenocarcinoma (LUAD), and colorectal carcinoma (CRC), highlighting the critical impact of comorbidity trajectories and treatment histories on outcomes. 
This analysis underscores the necessity of integrating heterogeneous multimodal data tailored to the unique characteristics of cancers, which can be enhanced by MFM with effective multimodal data integration.

We also implemented saliency mapping to identify morphometric, molecular, and clinical biomarkers.
As the BRCA samples shown in Figure \ref{fig:visual}c-e, we illustrate crucial factors in each modality.

For pathology images (Figure \ref{fig:visual}c), salient regions concentrated in tumor cores and invasive margins. 
To intuitively show the difference between two samples, we employed the Hover-Net \cite{graham2019hover} to identify the cell types in high-attention patches. 
We noticed that patients with poor prognosis exhibited more aggressive, infiltrative growth patterns, characterized by interdigitating tumor margins, and also showed a significantly higher prevalence of tumor necrosis.

To identify critical pathways associated with breast cancer prognosis, we extracted the top 5\% most important genes from the genomics branch and performed Gene Ontology (GO) and Kyoto Encyclopedia of Genes and Genomes (KEGG) enrichment analyses in Figure \ref{fig:visual}d.
The most enriched terms in BP (green bars) were developmental processes, including pattern specification process (44 genes, -log10(p.adjust) = 3.8) and regionalization (39 genes, -log10(p.adjust) = 3.2), potentially implicating dysregulated developmental pathways in tumor heterogeneity and progression \cite{li2010ratio}.
Moreover, neoron-related factors, such as synaptic membrane (33 genes, -log10(p.adjust) = 2.6), postsynaptic membrane (20 genes, -log10(p.adjust) = 2.2), neurotransmitter receptor activity (14 genes, -log10(p.adjust) = 1.8), and neuroactive ligand-receptor interaction (30 genes, -log10(p.adjust) = 2.5) suggesting the underlying mechanisms involve neuronal signaling and synaptic communication \cite{li2023exploring}.
In addition, the circadian entrainment (12 genes, -log10(p.adjust) = 1.6) in KEGG (blue bars) links disrupted circadian rhythm genes to metabolic reprogramming and therapy resistance \cite{d2023circadian, stevens2014breast}.
Collectively, these findings highlight dysregulated developmental programs, aberrant neuronal signaling, and disrupted circadian rhythms as key interconnected pathways influencing breast cancer prognosis.


For clinical reports, we generate a word cloud (Figure \ref{fig:visual}e) in which the sizes of the words represent their respective significance within MICE in predicting breast cancer survival outcomes.
Key terms such as immunotherapy, hysterectomy, and oophorectomy underscore the pivotal roles of treatment interventions in shaping these outcomes.
The presence of bronchopneumonia and atherosclerosis highlights the influence of systemic inflammation and vascular issues on both therapeutic tolerance and the risk of metastasis.
Additionally, terms like ecchymosis, lymphadenitis, and gynecomastia emerge as visible biomarkers signaling cancer progression.
This analysis emphasizes the diverse factors influencing prognosis, encompassing treatment responses, pathological characteristics, and concurrent health conditions, thereby revealing opportunities to enhance NLP pipelines for mining clinical text effectively.

\section{Discussion}
Multimodal data integration represents a transformative approach for comprehensive characterization of the tumor microenvironment (TME). 
Although multimodal AI models have shown promise in improving prognosis accuracy \cite{zhou2024cohort, zhou2023cross, Xu_2023_ICCV, zhang2024prototypical}, their clinical translation remains limited due to insufficient generalizability and inconsistent performance across diverse cohorts. 
A critical challenge is their frequent performance degradation when applied in independent cohorts with varying clinical characteristics, underscoring the urgent need for more effective and scalable multimodal data integration strategies. 
Foundation models (FMs), trained on large-scale datasets, offer a promising paradigm to overcome this issue, as evidenced by recent advances in modality-specific FMs that enhance generalizability in precision oncology \cite{yang2025foundation, ma2024towards, chen2024uni}.

However, developing multimodal foundation models (MFMs) for effective multimodal data integration faces several critical challenges. 
Most current approaches \cite{wang2024pathology, huang2023visual, xu2024multimodal} rely heavily on contrastive learning to align cross-modal representations—an approach that emphasizes shared information while often neglecting modality-specific insights.
Given the substantial heterogeneity between data modalities, such as genomics and imaging, this strategy tends to underutilize unique modal attributes and deeper inter-modal interactions. 
Moreover, the integration of multiple unimodel features is typically learned from limited samples in downstream tasks. 
It restricts the model’s ability to capture complex multimodal relationships essential for a holistic understanding of the TME and limits overall predictive performance. 
Alternative strategies, such as supervised pre-training using survival follow-up data \cite{keyl2025decoding}, have been proposed.
However, they simply extracted several pre-defined indicators from individual modalities like genomics and pathology images, potentially constraining the discovery of novel biomarkers and limiting overall predictive power.

To address these challenges, we developed MICE, to our knowledge, the first MFM designed for effective integration of genomics data, pathology images, and clinical reports. 
Built upon multimodal data from 11,799 patients across 30 cancer types, MICE effectively overcomes the scalability limitations of existing multimodal AI approaches.
Crucially, MICE employs a novel collaborative multi-expert module and a hybrid pre-training strategy, which together enhance both discriminative power and generalizability. 
By simultaneously capturing modality-shared and modality-specific patterns within a unified representation, MICE significantly improves modeling of complex tumor microenvironment (TME) interactions compared to current state-of-the-art models.

MICE demonstrated robust performance across diverse pan-cancer prognosis prediction tasks. 
Its novel integration strategy achieved substantial improvements in prognostic accuracy, with C-index increases ranging from 3.8\% to 11.2\% compared to existing AI models in internal cohorts.
Notably, MICE maintained comparable predictive precision against state-of-the-art multimodal AI models even when trained with only 50\% of the samples, demonstrating exceptional data efficiency—a key advantage in clinical settings where multimodal data are often limited and costly to acquire.
The model also consistently stratified patients into distinct risk groups, highlighting its potential clinical utility for guiding personalized treatment strategies and postoperative management. 
Together, these results underscore the strong generalizability of MICE across cancer types and patient populations, supporting its translational promise in bridging the gap between AI innovation and clinical application in precision oncology.

Despite these advances, several limitations should be noted:
1) \textit{Limited data scale and diversity.} 
Although The Cancer Genome Atlas (TCGA) was used as a primary pre-training resource, its multimodal sample size (approximately 10,000 patients) remains moderate by contemporary standards—for instance, compared to resources such as MIMIC-IV \cite{johnson2023mimic}, which contains hundreds of thousands of samples.
In addition, large-scale multimodal datasets suitable for extensive independent validation are still scarce. 
Future work should prioritize the collection of more diverse and larger multimodal cohorts to thoroughly evaluate the robustness and generalizability of multimodal AI models.
2) \textit{Architectural simplicity of the expert module.} 
While the expert-based design in MICE is effective, it offers room for architectural refinement. 
Developing more sophisticated modules that can efficiently learn fine-grained cross-modal interactions and explicitly model complex multimodal dynamics would enhance the representation of TME heterogeneity.
3) \textit{Modality coverage.} 
The current model integrates three prevalent data modalities, including pathology image, clinical report, and genomics data.
Inclusion of additional modalities (e.g., radiology images, demographic factors) could broaden clinical applicability and support a more holistic assessment of individual patient profiles.

In summary, MICE represents a paradigm shift in AI-driven oncology by integrating molecular, histopathological, and clinical data into a unified and generalizable MFM. 
It addresses the key barriers of the generalizability and effectiveness of multimodal data integration that have impeded previous approaches, offering a robust and scalable framework for clinically applicable multimodal AI.
Future efforts aimed at expanding data diversity, refining architectural design, and incorporating additional modalities will further accelerate the translation of multimodal foundation models such as MICE into routine clinical use, ultimately facilitating more personalized cancer therapy and improving treatment outcomes in precision oncology.

\section{Methods}
\subsection{Multimodal data collection}
To assemble the multimodal data, we collected 30 cohorts from the public TCGA dataset for developing and internally validating the MICE model.
For independent validation, we leveraged one public cohort from the HANCOCK dataset \cite{dorrich2025multimodal} and two in-house cohorts collected from three collaborating hospitals.
Specifically, in the in-house cohorts, we collected the hematoxylin and eosin (H\&E)-stained whole slide images as pathology data and corresponding pathology reports as clinical reports.
We also collected available follow-ups for prognosis prediction, including overall survival (OS), disease specific survival (DSS), progression-free survival (PFS), disease-free survival (DFS), local recurrence-free survival (LRFS), and distant metastasis-free survival (DMFS).
OS is defined as the time from diagnosis or treatment to death caused by any reason, while DSS is specific to the death caused by the target cancer.
PFS is the length of time during and after the treatment of a cancer, that a patient lives with the disease but it does not get worse.
DFS is the length of time after primary treatment for a cancer ends that the patient survives without any signs or symptoms of that cancer.
LRFS and DMFS indicates the the length of time after treatment that the patient survives without any local recurrence and distant metastasis of that cancer.

The study was reviewed and received approval by the Hong Kong University of Science and Technology (HREP-2025-0177).

\subsection{Unimodal feature extraction}
We adopted modality-specific approaches to extract features from each modality.
1) When processing pathology images, we first load the slides at the 20x magnification and subsequently apply the histogram-based Otsu segmentation method \cite{otsu1975threshold} to generate a binary mask to distinguish the foreground and background regions. 
Then, we partition the foreground regions into non-overlapping 512x512 patches and employ the pathology foundation model UNI \cite{chen2024uni} to derive 1024-dimensional patch embeddings, forming an embedding group for each WSI. 
In cases where a patient has multiple WSIs, multiple embedding groups are merged into a unified group.
In addition, we employ the MambaMIL method \cite{yang2024mambamil} to aggregate the embedding group into a patient-level representation.
2) Regarding genomics data, after z-score normalization of expression levels, we consolidate the gene sets from all TCGA cohorts to form a union gene ensemble. 
Genes absent from a patient’s profile were imputed with zeros, resulting in a total of 20,245 gene expression data points for all patients in the TCGA dataset.
To model the correlations between genes, we employ Self-Normalizing Networks (SNNs) \cite{klambauer2017self} for representation learning.
It should be noted that the genomics data used in this study is mRNA expression expression data, which are incompatible with existing genomics FMs \cite{dalla2025nucleotide} designed to operate on raw nucleotide sequences (ACGT strings).
3) In terms of clinical reports, we employ BioBERT \cite{lee2020biobert} to capture rich semantic information about tumor characteristics in clinical reports.
The majority of token sequences in the collected reports contained fewer than 512 tokens, which is the maximum length supported by BioBERT. 
For reports exceeding this limitation, we applied random truncation to ensure that they comply with the maximum length requirement for processing.
4) Alongside multimodal data from cancer patients, learnable cancer embeddings were provided to MICE, indicating the cancer types of patients.
These embeddings adaptively capture cancer-specific biological mechanisms, ensuring multimodal data integration varies for each cancer type.
For samples with missing modalities, we adapted the approach proposed by Liu et al. \cite{liu2023m3ae} to harness SNNs to synthesize the unimodal representation of missing modality directly from cancer embeddings. 
Therefore, MICE is capable of handling clinical scenarios where patients have incomplete multimodal data, enhancing its scalability to various clinical scenarios.

\begin{figure*}[!t]
\centering
    \includegraphics[width=0.99 \textwidth]{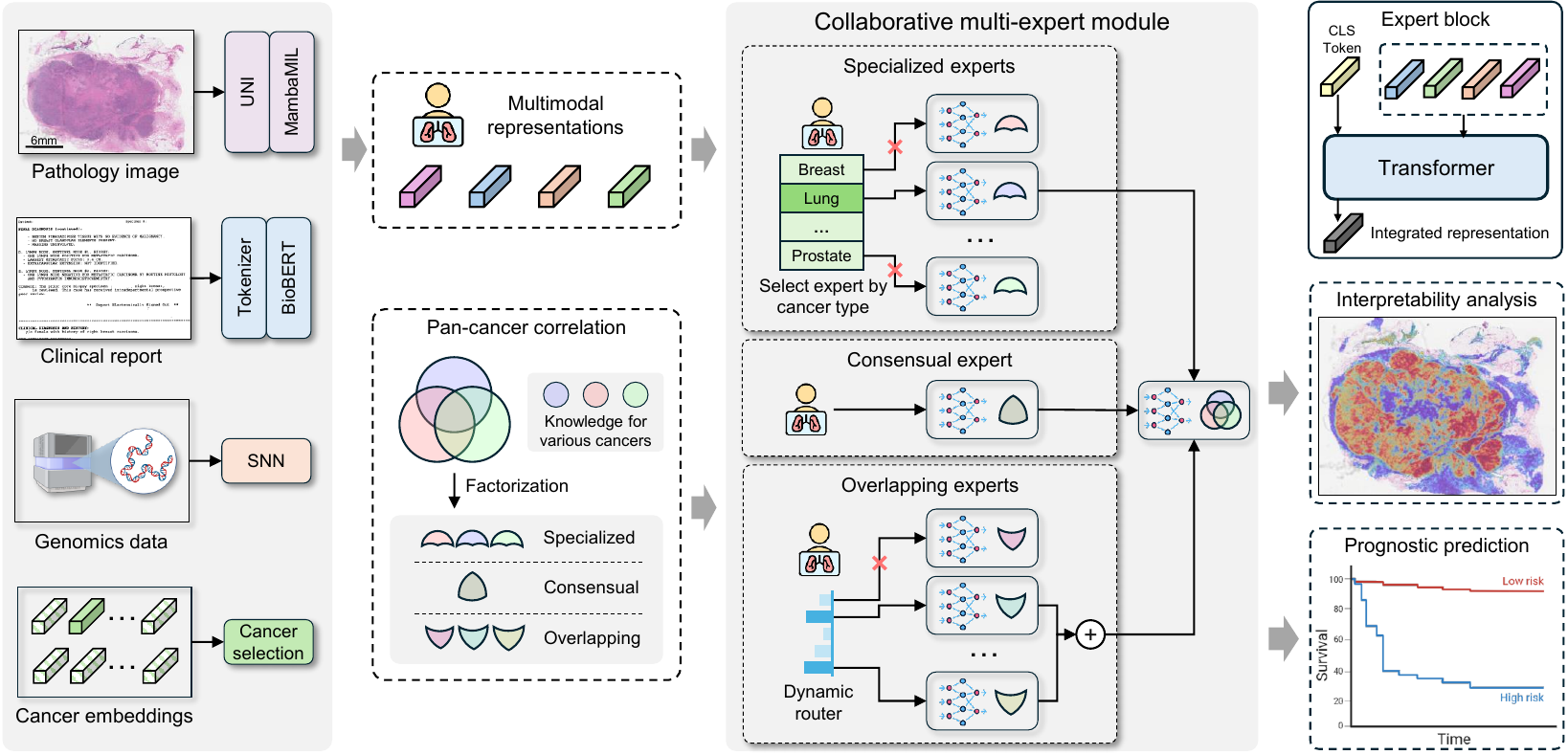}
\caption{\textbf{Model structure of MICE.} 
We first employ different approaches to extract unimodal representations.
Subsequently, we develop the collaborative multi-expert module to comprehensively model pan-cancer correlations and therefore enable effective and generalized multimodal data integration. Finally, an additional fusion expert is employed to produce patient-level prognosis prediction. All expert modules are implemented as Transformer networks.  
}
\label{fig:network}
\end{figure*}

\subsection{Collaborative multi-expert module for effective and generalized multimodal data integration}
Existing multimodal AI models for precision oncology face two crucial challenges, i.e., ineffective integration of multimodal data and poor generalizability.
In this study, we proposed MICE (Multimodal data Integration via Collaborative Experts) to address these challenges by leveraging a collaborative multi-expert module and a coupled pre-training strategy.

Two typical models capable of leveraging pan-cancer data for developing MFM are multi-head \cite{shao2024multi} and Mixture-of-Experts (MoE) \cite{xiong2024mome, zhou2022mixture, jiang2024m4oe} models.
Specifically, the multi-head model employs a single network to extract multimodal features from inputs features, while multiple cancer-specific expert modules are employed to produce prediction conditioned on cancer types.
The single feature extraction network is hard to capture cancer-specific details crucial for predicting the prognosis.
On the other hand, MoE model can mitigate this issue by introducing a routing mechanism that dynamically selects expert modules for feature learning.
However, this mechanism makes it difficult to capture cross-cancer consensus within any single expert module.
Furthermore, multiple expert modules employed in these models are functionally similar, limiting their ability in comprehensively capture distinct knowledge between cancers.  

To address this issue, we proposed a collaborative multi-expert module to comprehensively characterize the correlation between cancers, as illustrated in Figure \ref{fig:network}.
To exploit inter-cancer correlations, we organized multiple expert modules into three functionally distinct categories:
1) a consensual expert for identifying pan-cancer biological knowledge shared across all cancers;
2) a group of specialized experts for encoding unique characteristics for individual cancer types, similar to multiple classifiers in multi-head model;
3) a group of overlapping experts for dynamically acquiring the information shared among parts of cancers.
A router module is employed to decide which overlapping experts are selected to integrate multimodal features, similar to the MoE model.
During training, for a given sample, only the consensual expert, the specialized expert for corresponding cancer type, and the selected overlapping experts are employed for integrating multimodal inputs.
This mechanism inherently ensure that expert modules learn to capture distinct knowledge within pan-cancer data, shaping them as knowledge-specific experts.
Collectively, three expert categories can learn the consensual, overlapping, and specialized knowledge between cancers, forming the whole knowledge space of pan-cancer data.
Furthermore, consensual and overlapping experts are trained on samples of diverse cancer types, further enhancing the generalizability of the extracted features.
Our multi-expert module can maximize the advantage of pan-cancer data by comprehensively capture knowledge with prognosis significance, thereby facilitating effective multimodal data integration and improved prognosis accuracy.
Subsequently, the outputs of these expert categories are aggregated into a holistic patient-level feature via an additional fusion expert module. 
Finally, a classifier layer produces patient-level predictions from the fused feature.
All experts are implemented as Transformer networks \cite{vaswani2017attention} with multi-head self-attention to enable dynamic cross-modal interactions.

To enhance the generalizability, we couples self-supervised and supervised learning strategies to pre-train MICE.
For self-supervised learning, we employ contrastive learning \cite{chen2020simple} that aligns multimodal features of the same patient while enlarging the distances between features from different patients, thus helping extract representative features from each modality.
The formulation of contrastive learning can be written as:
\begin{equation}
    l_{cl} = - \log \frac{\sum_{{M}_{+}} d(M, {M}_{+})}{\sum_{{M}_{+}} d(M, {M}_{+}) + \sum_{{M}_{-}} d(M, {M}_{-})},
\end{equation}
where $M_+$ and $M_-$ indicates the positive and negative features, respectively.
$d$ calculates the similarity between two inputs. 
For supervised learning, prognosis follow-ups are used to supervise the multimodal data integration process, guiding MICE to better learn the correlations between cancers.
We generalize the negative log-likelihood (NLL) with censorship to supervise the prognosis prediction \cite{zhou2024cohort, zhou2023cross, Xu_2023_ICCV} by:
\begin{equation}
    \begin{split}
    L_{surv}=&-c\log(f_{sur}(H,k))\\
    &-(1-c)\log(f_{sur}(H,{k-1}))\\
    &-(1-c)\log(h_k).
    \end{split}
\end{equation}
Each patient sample is defined as a triplet $\{H, c, k\}$, where $H=\{h_1,...h_n\}$ is the predicted hazard vector that measures the probability of the time of the ground truth event located in the corresponding time interval $k$.
$f_{sur}(H,k)=\prod_{j=1}^{k}(1-h_{j})$ is the discrete survival function that calculates the probability of survival at time interval $k$.
$c$ is the censorship indicator, 0 for censored and 1 for uncensored.
The final loss for pretraining MICE is a combination of these two loss function:
\begin{equation}
    L=L_{cl}+ \alpha L_{surv},
\end{equation}
where $\alpha$ is a hyperparameter controlling the weights of losses. We set it to 2, which achieves the best performance in our internal validation experiments.

\subsection{Implementation details}
During implementation, our code was based on Python 3 and the open-source PyTorch library with an NVIDIA 3090 GPU equipped with 24GB memory.
For model development, MICE was pre-trained and internally validated on the TCGA dataset following five-fold cross-validation. 
It underwent a pre-training phase of 30 epochs followed by finetuning across 20 epochs on internal cohorts.
We employed Adam as the optimizer with a learning rate of $1e^{-4}$ for all model training processes.
For independent validation, we first pre-trained MICE on all samples in the TCGA dataset and finetuned the pre-trained model on independent cohorts via five-fold cross-validation.

\subsection{Compared models}
To demonstrate the effectiveness of MICE, we construct multiple models as the baselines for comparison, leveraging the same modality-specific encoders as our MICE.
First, three unimodal models utilize pathology images, clinical reports, and molecular profiles, separately. 
Second, an early fusion model that simply concatenates multiple unimodal representations and produce the prognosis prediction using an MLP classifier.
Third, a late fusion model that integrates multiple predictions generated from each modality alone.
Fourth, a multi-head network \cite{shao2024multi} that concatenates multimodal representations and employs multiple classifiers to produce the predictions for different cancer types.
Last, a Mixture-of-Experts (MoE) model \cite{jiang2024m4oe} that selects experts conditioned on the input data.
The selection process indicates that no expert will be selected for all samples, which makes it difficult to learn the consensual knowledge across all cancers.
For independent validation, since all independent cohorts only contains two modalities, we further compare MICE with two latest bimodal models, MCAT \cite{mcat} and SurvPath \cite{jaume2024modeling}, which are designed for integrating pathology images with molecular profiles.
For application to our independent cohorts, which include pathology images and clinical reports, we substitute the genomics features with text features derived from the same text encoder in MICE.

\subsection{Statistical analysis}
The assessment of prognosis prediction performance involves computing the C-index, which examines the pairwise concordance of predicted prognosis outcomes among patient pairs. 
The $p$-$value$ between two models is calculated using the two-sided Wilcoxon signed-rank test \cite{wilcoxon1992individual}.
In addition, Kaplan-Meier analysis was employed to reinforce the evaluation by delineating the significance of differences between the predicted high- and low-risk groups, while the $p$-$value$ is obtained by the log-rank test. 
We employed the KaplanMeierFitter and logrank\_test methods in the lifelines library for Kaplan-Meier analysis and corresponding $p$-$value$ calculation.

\subsection{Interpretability analysis}
We adopted various approaches for interpretability analysis on MICE.
First, to quantifying the contribution of each modality, we adopted the SHAP value estimation method presented in \cite{lundberg2017unified} as a unified measure of modality importance. 
Second, we adopted different approaches for intre-modality analysis.
For pathology image, we extracted the attention scores in MambaMIL \cite{yang2024mambamil} to build the attention maps for WSIs.
As for molecular profiles, we collected the average weights in the first SNN layer as the importance scores of genes.
For clinical reports, the attention vectors in the pre-trained BioBERT are employed to understand which words are crucial for prognosis prediction.

\section{Data availability}\label{sec4}
The in-house data used in this study is not publicly available due to compliance with patient privacy protection.
Corresponding author can be reached for data usage agreement for approving the access to the private data.
For TCGA dataset, pathology images, mRNA expression data, and a clean version of clinical reports are publicly available at \href{https://portal.gdc.cancer.gov/}{https://portal.gdc.cancer.gov/}, \href{https://www.cbioportal.org/}{https://www.cbioportal.org/}, and \href{https://github.com/cpystan/Wsi-Caption}{https://github.com/cpystan/Wsi-Caption}, respectively.
The HANCOCK dataset is publicly available at \href{https://hancock.research.fau.eu}{https://hancock.research.fau.eu}.

\section{Code availability}\label{sec12}
The source code for the MICE model presented in this study has been made available at: \href{https://github.com/moothes/MICE}{https://github.com/moothes/MICE}.

\section{Author contributions}
H.J.Z., F.T.Z., J.B.M., Y.X.X, X.W., and H.C. designed the model structure and pre-training strategy. 
J.B.M., Y.X.X., X.M.Z., L.L., and Z.H.L. were responsible for collecting in-house cohorts. 
H.J.Z., F.T.Z., J.B.M., and Y.X.X. implemented the pre-training and independent validation.
H.J.Z., F.T.Z., and X.W. co-wrote the manuscript. 
H.C. supervised this study and critically revised the manuscript. 
All authors contributed to the editing of the revised manuscript and approved the manuscript.

\section{Acknowledgements}
This work was supported by the National Natural Science Foundation of China (No. 62202403), Innovation and Technology Commission (Project No. MHP/002/22 and ITCPD/17-9), Research Grants Council of the Hong Kong Special Administrative Region, China (Project No: T45-401/22-N) and National Key R\&D Program of China (Project No. 2023YFE0204000).  


\section{Competing interests}
The authors declare no competing interests.

\backmatter


\bibliography{sn-bibliography}


\begin{thebibliography}{56}
\ifx \bisbn   \undefined \def \bisbn  #1{ISBN #1}\fi
\ifx \binits  \undefined \def \binits#1{#1}\fi
\ifx \bauthor  \undefined \def \bauthor#1{#1}\fi
\ifx \batitle  \undefined \def \batitle#1{#1}\fi
\ifx \bjtitle  \undefined \def \bjtitle#1{#1}\fi
\ifx \bvolume  \undefined \def \bvolume#1{\textbf{#1}}\fi
\ifx \byear  \undefined \def \byear#1{#1}\fi
\ifx \bissue  \undefined \def \bissue#1{#1}\fi
\ifx \bfpage  \undefined \def \bfpage#1{#1}\fi
\ifx \blpage  \undefined \def \blpage #1{#1}\fi
\ifx \burl  \undefined \def \burl#1{\textsf{#1}}\fi
\ifx \doiurl  \undefined \def \doiurl#1{\url{https://doi.org/#1}}\fi
\ifx \betal  \undefined \def \betal{\textit{et al.}}\fi
\ifx \binstitute  \undefined \def \binstitute#1{#1}\fi
\ifx \binstitutionaled  \undefined \def \binstitutionaled#1{#1}\fi
\ifx \bctitle  \undefined \def \bctitle#1{#1}\fi
\ifx \beditor  \undefined \def \beditor#1{#1}\fi
\ifx \bpublisher  \undefined \def \bpublisher#1{#1}\fi
\ifx \bbtitle  \undefined \def \bbtitle#1{#1}\fi
\ifx \bedition  \undefined \def \bedition#1{#1}\fi
\ifx \bseriesno  \undefined \def \bseriesno#1{#1}\fi
\ifx \blocation  \undefined \def \blocation#1{#1}\fi
\ifx \bsertitle  \undefined \def \bsertitle#1{#1}\fi
\ifx \bsnm \undefined \def \bsnm#1{#1}\fi
\ifx \bsuffix \undefined \def \bsuffix#1{#1}\fi
\ifx \bparticle \undefined \def \bparticle#1{#1}\fi
\ifx \barticle \undefined \def \barticle#1{#1}\fi
\bibcommenthead
\ifx \bconfdate \undefined \def \bconfdate #1{#1}\fi
\ifx \botherref \undefined \def \botherref #1{#1}\fi
\ifx \url \undefined \def \url#1{\textsf{#1}}\fi
\ifx \bchapter \undefined \def \bchapter#1{#1}\fi
\ifx \bbook \undefined \def \bbook#1{#1}\fi
\ifx \bcomment \undefined \def \bcomment#1{#1}\fi
\ifx \oauthor \undefined \def \oauthor#1{#1}\fi
\ifx \citeauthoryear \undefined \def \citeauthoryear#1{#1}\fi
\ifx \endbibitem  \undefined \def \endbibitem {}\fi
\ifx \bconflocation  \undefined \def \bconflocation#1{#1}\fi
\ifx \arxivurl  \undefined \def \arxivurl#1{\textsf{#1}}\fi
\csname PreBibitemsHook\endcsname

\bibitem[\protect\citeauthoryear{Bray et~al.}{2024}]{bray2024global}
\begin{barticle}
\bauthor{\bsnm{Bray}, \binits{F.}}, \betal:
\batitle{Global cancer statistics 2022: Globocan estimates of incidence and mortality worldwide for 36 cancers in 185 countries}.
\bjtitle{CA: a cancer journal for clinicians}
\bvolume{74}(\bissue{3}),
\bfpage{229}--\blpage{263}
(\byear{2024})
\end{barticle}
\endbibitem

\bibitem[\protect\citeauthoryear{Sparano et~al.}{2018}]{sparano2018adjuvant}
\begin{barticle}
\bauthor{\bsnm{Sparano}, \binits{J.A.}}, \betal:
\batitle{Adjuvant chemotherapy guided by a 21-gene expression assay in breast cancer}.
\bjtitle{New England Journal of Medicine}
\bvolume{379}(\bissue{2}),
\bfpage{111}--\blpage{121}
(\byear{2018})
\end{barticle}
\endbibitem

\bibitem[\protect\citeauthoryear{Hui et~al.}{2021}]{hui2021importance}
\begin{barticle}
\bauthor{\bsnm{Hui}, \binits{D.}},
\bauthor{\bsnm{Mo}, \binits{L.}},
\bauthor{\bsnm{Paiva}, \binits{C.E.}}:
\batitle{The importance of prognostication: impact of prognostic predictions, disclosures, awareness, and acceptance on patient outcomes}.
\bjtitle{Current treatment options in oncology}
\bvolume{22},
\bfpage{1}--\blpage{14}
(\byear{2021})
\end{barticle}
\endbibitem

\bibitem[\protect\citeauthoryear{Orlovic et~al.}{2023}]{orlovic2023accuracy}
\begin{barticle}
\bauthor{\bsnm{Orlovic}, \binits{M.}}, \betal:
\batitle{Accuracy of clinical predictions of prognosis at the end-of-life: evidence from routinely collected data in urgent care records}.
\bjtitle{BMC Palliative Care}
\bvolume{22}(\bissue{1}),
\bfpage{51}
(\byear{2023})
\end{barticle}
\endbibitem

\bibitem[\protect\citeauthoryear{Boehm et~al.}{2022}]{boehm2022multimodal}
\begin{barticle}
\bauthor{\bsnm{Boehm}, \binits{K.M.}}, \betal:
\batitle{Multimodal data integration using machine learning improves risk stratification of high-grade serous ovarian cancer}.
\bjtitle{Nature cancer}
\bvolume{3}(\bissue{6}),
\bfpage{723}--\blpage{733}
(\byear{2022})
\end{barticle}
\endbibitem

\bibitem[\protect\citeauthoryear{Bortolini~Silveira et~al.}{2021}]{bortolini2021multimodal}
\begin{barticle}
\bauthor{\bsnm{Bortolini~Silveira}, \binits{A.}}, \betal:
\batitle{Multimodal liquid biopsy for early monitoring and outcome prediction of chemotherapy in metastatic breast cancer}.
\bjtitle{NPJ breast cancer}
\bvolume{7}(\bissue{1}),
\bfpage{115}
(\byear{2021})
\end{barticle}
\endbibitem

\bibitem[\protect\citeauthoryear{Gao et~al.}{2024}]{gao2024explainable}
\begin{barticle}
\bauthor{\bsnm{Gao}, \binits{Y.}}, \betal:
\batitle{An explainable longitudinal multi-modal fusion model for predicting neoadjuvant therapy response in women with breast cancer}.
\bjtitle{Nature Communications}
\bvolume{15}(\bissue{1}),
\bfpage{9613}
(\byear{2024})
\end{barticle}
\endbibitem

\bibitem[\protect\citeauthoryear{Zhou et~al.}{2024}]{zhou2024multimodal}
\begin{botherref}
\oauthor{\bsnm{Zhou}, \binits{H.}}, et al.:
Multimodal data integration for precision oncology: Challenges and future directions.
arXiv preprint arXiv:2406.19611
(2024)
\end{botherref}
\endbibitem

\bibitem[\protect\citeauthoryear{Nakach et~al.}{2024}]{nakach2024comprehensive}
\begin{barticle}
\bauthor{\bsnm{Nakach}, \binits{F.-Z.}},
\bauthor{\bsnm{Idri}, \binits{A.}},
\bauthor{\bsnm{Goceri}, \binits{E.}}:
\batitle{A comprehensive investigation of multimodal deep learning fusion strategies for breast cancer classification}.
\bjtitle{Artificial Intelligence Review}
\bvolume{57}(\bissue{12}),
\bfpage{327}
(\byear{2024})
\end{barticle}
\endbibitem

\bibitem[\protect\citeauthoryear{Boehm et~al.}{2022}]{boehm2022harnessing}
\begin{barticle}
\bauthor{\bsnm{Boehm}, \binits{K.M.}},
\bauthor{\bsnm{Khosravi}, \binits{P.}},
\bauthor{\bsnm{Vanguri}, \binits{R.}},
\bauthor{\bsnm{Gao}, \binits{J.}},
\bauthor{\bsnm{Shah}, \binits{S.P.}}:
\batitle{Harnessing multimodal data integration to advance precision oncology}.
\bjtitle{Nature Reviews Cancer}
\bvolume{22}(\bissue{2}),
\bfpage{114}--\blpage{126}
(\byear{2022})
\end{barticle}
\endbibitem

\bibitem[\protect\citeauthoryear{Zhang et~al.}{2024}]{zhang2024prototypical}
\begin{bchapter}
\bauthor{\bsnm{Zhang}, \binits{Y.}},
\bauthor{\bsnm{Xu}, \binits{Y.}},
\bauthor{\bsnm{Chen}, \binits{J.}},
\bauthor{\bsnm{Xie}, \binits{F.}},
\bauthor{\bsnm{Chen}, \binits{H.}}:
\bctitle{Prototypical information bottlenecking and disentangling for multimodal cancer survival prediction}.
(\byear{2024})
\end{bchapter}
\endbibitem

\bibitem[\protect\citeauthoryear{Byeon et~al.}{2025}]{byeon2025interpretable}
\begin{barticle}
\bauthor{\bsnm{Byeon}, \binits{Y.}}, \betal:
\batitle{Interpretable multimodal transformer for prediction of molecular subtypes and grades in adult-type diffuse gliomas}.
\bjtitle{NPJ Digital Medicine}
\bvolume{8}(\bissue{1}),
\bfpage{140}
(\byear{2025})
\end{barticle}
\endbibitem

\bibitem[\protect\citeauthoryear{Song et~al.}{2024}]{song2024multimodal}
\begin{bchapter}
\bauthor{\bsnm{Song}, \binits{A.H.}}, \betal:
\bctitle{Multimodal prototyping for cancer survival prediction}.
In: \bbtitle{Forty-first International Conference on Machine Learning}
(\byear{2024})
\end{bchapter}
\endbibitem

\bibitem[\protect\citeauthoryear{Volinsky-Fremond et~al.}{2024}]{volinsky2024prediction}
\begin{barticle}
\bauthor{\bsnm{Volinsky-Fremond}, \binits{S.}}, \betal:
\batitle{Prediction of recurrence risk in endometrial cancer with multimodal deep learning}.
\bjtitle{Nature Medicine}
\bvolume{30}(\bissue{7}),
\bfpage{1962}--\blpage{1973}
(\byear{2024})
\end{barticle}
\endbibitem

\bibitem[\protect\citeauthoryear{Xiang et~al.}{2024}]{xiang2024development}
\begin{barticle}
\bauthor{\bsnm{Xiang}, \binits{H.}}, \betal:
\batitle{Development and validation of an interpretable model integrating multimodal information for improving ovarian cancer diagnosis}.
\bjtitle{Nature Communications}
\bvolume{15}(\bissue{1}),
\bfpage{2681}
(\byear{2024})
\end{barticle}
\endbibitem

\bibitem[\protect\citeauthoryear{Xiong et~al.}{2024}]{xiong2024mome}
\begin{bchapter}
\bauthor{\bsnm{Xiong}, \binits{C.}}, \betal:
\bctitle{Mome: Mixture of multimodal experts for cancer survival prediction}.
In: \bbtitle{International Conference on Medical Image Computing and Computer-Assisted Intervention},
pp. \bfpage{318}--\blpage{328}
(\byear{2024}).
\bcomment{Springer}
\end{bchapter}
\endbibitem

\bibitem[\protect\citeauthoryear{Qian et~al.}{2024}]{qian2024multimodal}
\begin{botherref}
\oauthor{\bsnm{Qian}, \binits{X.}}, et al.:
A multimodal machine learning model for the stratification of breast cancer risk.
Nature Biomedical Engineering,
1--15
(2024)
\end{botherref}
\endbibitem

\bibitem[\protect\citeauthoryear{Sharma et~al.}{2025}]{sharma2025hybrid}
\begin{barticle}
\bauthor{\bsnm{Sharma}, \binits{P.}}, \betal:
\batitle{A hybrid multi model artificial intelligence approach for glaucoma screening using fundus images}.
\bjtitle{npj Digital Medicine}
\bvolume{8}(\bissue{1}),
\bfpage{130}
(\byear{2025})
\end{barticle}
\endbibitem

\bibitem[\protect\citeauthoryear{Bai et~al.}{2025}]{bai2025predicting}
\begin{barticle}
\bauthor{\bsnm{Bai}, \binits{Z.}}, \betal:
\batitle{Predicting response to neoadjuvant chemotherapy in muscle-invasive bladder cancer via interpretable multimodal deep learning}.
\bjtitle{npj Digital Medicine}
\bvolume{8}(\bissue{1}),
\bfpage{174}
(\byear{2025})
\end{barticle}
\endbibitem

\bibitem[\protect\citeauthoryear{Xu et~al.}{2024}]{xu2024multimodal}
\begin{botherref}
\oauthor{\bsnm{Xu}, \binits{Y.}}, et al.:
A multimodal knowledge-enhanced whole-slide pathology foundation model.
arXiv preprint arXiv:2407.15362
(2024)
\end{botherref}
\endbibitem

\bibitem[\protect\citeauthoryear{Lu et~al.}{2024}]{lu2024visual}
\begin{barticle}
\bauthor{\bsnm{Lu}, \binits{M.Y.}}, \betal:
\batitle{A visual-language foundation model for computational pathology}.
\bjtitle{Nature Medicine}
\bvolume{30}(\bissue{3}),
\bfpage{863}--\blpage{874}
(\byear{2024})
\end{barticle}
\endbibitem

\bibitem[\protect\citeauthoryear{Xiang et~al.}{2025}]{xiang2025vision}
\begin{botherref}
\oauthor{\bsnm{Xiang}, \binits{J.}}, et al.:
A vision--language foundation model for precision oncology.
Nature,
1--10
(2025)
\end{botherref}
\endbibitem

\bibitem[\protect\citeauthoryear{Huang et~al.}{2023}]{huang2023visual}
\begin{botherref}
\oauthor{\bsnm{Huang}, \binits{Z.}},
\oauthor{\bsnm{Bianchi}, \binits{F.}},
\oauthor{\bsnm{Yuksekgonul}, \binits{M.}},
\oauthor{\bsnm{Montine}, \binits{T.J.}},
\oauthor{\bsnm{Zou}, \binits{J.}}:
A visual--language foundation model for pathology image analysis using medical twitter.
Nature Medicine,
1--10
(2023)
\end{botherref}
\endbibitem

\bibitem[\protect\citeauthoryear{Wang et~al.}{2023}]{wang2023shared}
\begin{barticle}
\bauthor{\bsnm{Wang}, \binits{Z.}},
\bauthor{\bsnm{Yu}, \binits{L.}},
\bauthor{\bsnm{Ding}, \binits{X.}},
\bauthor{\bsnm{Liao}, \binits{X.}},
\bauthor{\bsnm{Wang}, \binits{L.}}:
\batitle{Shared-specific feature learning with bottleneck fusion transformer for multi-modal whole slide image analysis}.
\bjtitle{IEEE Transactions on Medical Imaging}
\bvolume{42}(\bissue{11}),
\bfpage{3374}--\blpage{3383}
(\byear{2023})
\end{barticle}
\endbibitem

\bibitem[\protect\citeauthoryear{Shao et~al.}{2023}]{shao2023fam3l}
\begin{barticle}
\bauthor{\bsnm{Shao}, \binits{W.}}, \betal:
\batitle{Fam3l: Feature-aware multi-modal metric learning for integrative survival analysis of human cancers}.
\bjtitle{IEEE Transactions on Medical Imaging}
\bvolume{42}(\bissue{9}),
\bfpage{2552}--\blpage{2565}
(\byear{2023})
\end{barticle}
\endbibitem

\bibitem[\protect\citeauthoryear{Shao et~al.}{2024}]{shao2024multi}
\begin{barticle}
\bauthor{\bsnm{Shao}, \binits{W.}},
\bauthor{\bsnm{Shi}, \binits{H.}},
\bauthor{\bsnm{Liu}, \binits{J.}},
\bauthor{\bsnm{Zuo}, \binits{Y.}},
\bauthor{\bsnm{Sun}, \binits{L.}},
\bauthor{\bsnm{Xia}, \binits{T.}},
\bauthor{\bsnm{Chen}, \binits{W.}},
\bauthor{\bsnm{Wan}, \binits{P.}},
\bauthor{\bsnm{Sheng}, \binits{J.}},
\bauthor{\bsnm{Zhu}, \binits{Q.}}, \betal:
\batitle{Multi-instance multi-task learning for joint clinical outcome and genomic profile predictions from the histopathological images}.
\bjtitle{IEEE transactions on medical imaging}
\bvolume{43}(\bissue{6}),
\bfpage{2266}--\blpage{2278}
(\byear{2024})
\end{barticle}
\endbibitem

\bibitem[\protect\citeauthoryear{Zhou et~al.}{2022}]{zhou2022mixture}
\begin{barticle}
\bauthor{\bsnm{Zhou}, \binits{Y.}}, \betal:
\batitle{Mixture-of-experts with expert choice routing}.
\bjtitle{Advances in Neural Information Processing Systems}
\bvolume{35},
\bfpage{7103}--\blpage{7114}
(\byear{2022})
\end{barticle}
\endbibitem

\bibitem[\protect\citeauthoryear{Jiang and Shen}{2024}]{jiang2024m4oe}
\begin{bchapter}
\bauthor{\bsnm{Jiang}, \binits{Y.}},
\bauthor{\bsnm{Shen}, \binits{Y.}}:
\bctitle{M4oe: A foundation model for medical multimodal image segmentation with mixture of experts}.
In: \bbtitle{International Conference on Medical Image Computing and Computer-assisted Intervention},
pp. \bfpage{621}--\blpage{631}
(\byear{2024}).
\bcomment{Springer}
\end{bchapter}
\endbibitem

\bibitem[\protect\citeauthoryear{D{\"o}rrich et~al.}{2024}]{doerrich2024multimodal}
\begin{botherref}
\oauthor{\bsnm{D{\"o}rrich}, \binits{M.}}, et al.:
A multimodal dataset for precision oncology in head and neck cancer.
medRxiv,
2024--05
(2024)
\end{botherref}
\endbibitem

\bibitem[\protect\citeauthoryear{Chen et~al.}{2021}]{mcat}
\begin{bchapter}
\bauthor{\bsnm{Chen}, \binits{R.J.}}, \betal:
\bctitle{Multimodal co-attention transformer for survival prediction in gigapixel whole slide images}.
In: \bbtitle{Proceedings of the IEEE/CVF International Conference on Computer Vision},
pp. \bfpage{4015}--\blpage{4025}
(\byear{2021})
\end{bchapter}
\endbibitem

\bibitem[\protect\citeauthoryear{Jaume et~al.}{2024}]{jaume2024modeling}
\begin{bchapter}
\bauthor{\bsnm{Jaume}, \binits{G.}}, \betal:
\bctitle{Modeling dense multimodal interactions between biological pathways and histology for survival prediction}.
In: \bbtitle{Proceedings of the IEEE/CVF Conference on Computer Vision and Pattern Recognition},
pp. \bfpage{11579}--\blpage{11590}
(\byear{2024})
\end{bchapter}
\endbibitem

\bibitem[\protect\citeauthoryear{Graham et~al.}{2019}]{graham2019hover}
\begin{botherref}
\oauthor{\bsnm{Graham}, \binits{S.}}, et al.:
Hover-net: Simultaneous segmentation and classification of nuclei in multi-tissue histology images.
Medical Image Analysis,
101563
(2019)
\end{botherref}
\endbibitem

\bibitem[\protect\citeauthoryear{Li et~al.}{2010}]{li2010ratio}
\begin{barticle}
\bauthor{\bsnm{Li}, \binits{Y.}},
\bauthor{\bsnm{Zhang}, \binits{M.}},
\bauthor{\bsnm{Chen}, \binits{H.}},
\bauthor{\bsnm{Dong}, \binits{Z.}},
\bauthor{\bsnm{Ganapathy}, \binits{V.}},
\bauthor{\bsnm{Thangaraju}, \binits{M.}},
\bauthor{\bsnm{Huang}, \binits{S.}}:
\batitle{Ratio of mir-196s to hoxc8 messenger rna correlates with breast cancer cell migration and metastasis}.
\bjtitle{Cancer research}
\bvolume{70}(\bissue{20}),
\bfpage{7894}--\blpage{7904}
(\byear{2010})
\end{barticle}
\endbibitem

\bibitem[\protect\citeauthoryear{Li et~al.}{2023}]{li2023exploring}
\begin{barticle}
\bauthor{\bsnm{Li}, \binits{R.Q.}},
\bauthor{\bsnm{Zhao}, \binits{X.H.}},
\bauthor{\bsnm{Zhu}, \binits{Q.}},
\bauthor{\bsnm{Liu}, \binits{T.}},
\bauthor{\bsnm{Hondermarck}, \binits{H.}},
\bauthor{\bsnm{Thorne}, \binits{R.F.}},
\bauthor{\bsnm{Zhang}, \binits{X.D.}},
\bauthor{\bsnm{Gao}, \binits{J.N.}}:
\batitle{Exploring neurotransmitters and their receptors for breast cancer prevention and treatment}.
\bjtitle{Theranostics}
\bvolume{13}(\bissue{3}),
\bfpage{1109}
(\byear{2023})
\end{barticle}
\endbibitem

\bibitem[\protect\citeauthoryear{D’cunha et~al.}{2023}]{d2023circadian}
\begin{barticle}
\bauthor{\bsnm{D’cunha}, \binits{K.}},
\bauthor{\bsnm{Park}, \binits{Y.}},
\bauthor{\bsnm{Protani}, \binits{M.M.}},
\bauthor{\bsnm{Reeves}, \binits{M.M.}}:
\batitle{Circadian rhythm disrupting behaviours and cancer outcomes in breast cancer survivors: a systematic review}.
\bjtitle{Breast Cancer Research and Treatment}
\bvolume{198}(\bissue{3}),
\bfpage{413}--\blpage{421}
(\byear{2023})
\end{barticle}
\endbibitem

\bibitem[\protect\citeauthoryear{Stevens et~al.}{2014}]{stevens2014breast}
\begin{barticle}
\bauthor{\bsnm{Stevens}, \binits{R.G.}},
\bauthor{\bsnm{Brainard}, \binits{G.C.}},
\bauthor{\bsnm{Blask}, \binits{D.E.}},
\bauthor{\bsnm{Lockley}, \binits{S.W.}},
\bauthor{\bsnm{Motta}, \binits{M.E.}}:
\batitle{Breast cancer and circadian disruption from electric lighting in the modern world}.
\bjtitle{CA: a cancer journal for clinicians}
\bvolume{64}(\bissue{3}),
\bfpage{207}--\blpage{218}
(\byear{2014})
\end{barticle}
\endbibitem

\bibitem[\protect\citeauthoryear{Zhou et~al.}{2025}]{zhou2024cohort}
\begin{barticle}
\bauthor{\bsnm{Zhou}, \binits{H.}},
\bauthor{\bsnm{Zhou}, \binits{F.}},
\bauthor{\bsnm{Chen}, \binits{H.}}:
\batitle{Cohort-individual cooperative learning for multimodal cancer survival analysis}.
\bjtitle{IEEE Transactions on Medical Imaging}
\bvolume{44}(\bissue{2}),
\bfpage{656}--\blpage{667}
(\byear{2025})
\end{barticle}
\endbibitem

\bibitem[\protect\citeauthoryear{Zhou and Chen}{2023}]{zhou2023cross}
\begin{bchapter}
\bauthor{\bsnm{Zhou}, \binits{F.}},
\bauthor{\bsnm{Chen}, \binits{H.}}:
\bctitle{Cross-modal translation and alignment for survival analysis}.
In: \bbtitle{Proceedings of the IEEE/CVF International Conference on Computer Vision},
pp. \bfpage{21485}--\blpage{21494}
(\byear{2023})
\end{bchapter}
\endbibitem

\bibitem[\protect\citeauthoryear{Xu and Chen}{2023}]{Xu_2023_ICCV}
\begin{bchapter}
\bauthor{\bsnm{Xu}, \binits{Y.}},
\bauthor{\bsnm{Chen}, \binits{H.}}:
\bctitle{Multimodal optimal transport-based co-attention transformer with global structure consistency for survival prediction}.
In: \bbtitle{Proceedings of the IEEE/CVF International Conference on Computer Vision},
pp. \bfpage{21241}--\blpage{21251}
(\byear{2023})
\end{bchapter}
\endbibitem

\bibitem[\protect\citeauthoryear{Yang et~al.}{2025}]{yang2025foundation}
\begin{barticle}
\bauthor{\bsnm{Yang}, \binits{Z.}}, \betal:
\batitle{A foundation model for generalizable cancer diagnosis and survival prediction from histopathological images}.
\bjtitle{Nature Communications}
\bvolume{16}(\bissue{1}),
\bfpage{2366}
(\byear{2025})
\end{barticle}
\endbibitem

\bibitem[\protect\citeauthoryear{Ma et~al.}{2024}]{ma2024towards}
\begin{botherref}
\oauthor{\bsnm{Ma}, \binits{J.}}, et al.:
Towards a generalizable pathology foundation model via unified knowledge distillation.
arXiv preprint arXiv:2407.18449
(2024)
\end{botherref}
\endbibitem

\bibitem[\protect\citeauthoryear{Chen et~al.}{2024}]{chen2024uni}
\begin{botherref}
\oauthor{\bsnm{Chen}, \binits{R.J.}}, et al.:
Towards a general-purpose foundation model for computational pathology.
Nature Medicine
(2024)
\end{botherref}
\endbibitem

\bibitem[\protect\citeauthoryear{Wang et~al.}{2024}]{wang2024pathology}
\begin{barticle}
\bauthor{\bsnm{Wang}, \binits{X.}}, \betal:
\batitle{A pathology foundation model for cancer diagnosis and prognosis prediction}.
\bjtitle{Nature}
\bvolume{634}(\bissue{8035}),
\bfpage{970}--\blpage{978}
(\byear{2024})
\end{barticle}
\endbibitem

\bibitem[\protect\citeauthoryear{Keyl et~al.}{2025}]{keyl2025decoding}
\begin{botherref}
\oauthor{\bsnm{Keyl}, \binits{J.}}, et al.:
Decoding pan-cancer treatment outcomes using multimodal real-world data and explainable artificial intelligence.
Nature Cancer,
1--16
(2025)
\end{botherref}
\endbibitem

\bibitem[\protect\citeauthoryear{Johnson et~al.}{2023}]{johnson2023mimic}
\begin{barticle}
\bauthor{\bsnm{Johnson}, \binits{A.E.}}, \betal:
\batitle{Mimic-iv, a freely accessible electronic health record dataset}.
\bjtitle{Scientific data}
\bvolume{10}(\bissue{1}),
\bfpage{1}
(\byear{2023})
\end{barticle}
\endbibitem

\bibitem[\protect\citeauthoryear{D{\"o}rrich et~al.}{2025}]{dorrich2025multimodal}
\begin{barticle}
\bauthor{\bsnm{D{\"o}rrich}, \binits{M.}},
\bauthor{\bsnm{Balk}, \binits{M.}},
\bauthor{\bsnm{Heusinger}, \binits{T.}},
\bauthor{\bsnm{Beyer}, \binits{S.}},
\bauthor{\bsnm{Mirbagheri}, \binits{H.}},
\bauthor{\bsnm{Fischer}, \binits{D.J.}},
\bauthor{\bsnm{Kanso}, \binits{H.}},
\bauthor{\bsnm{Matek}, \binits{C.}},
\bauthor{\bsnm{Hartmann}, \binits{A.}},
\bauthor{\bsnm{Iro}, \binits{H.}}, \betal:
\batitle{A multimodal dataset for precision oncology in head and neck cancer}.
\bjtitle{Nature Communications}
\bvolume{16}(\bissue{1}),
\bfpage{7163}
(\byear{2025})
\end{barticle}
\endbibitem

\bibitem[\protect\citeauthoryear{Otsu et~al.}{1975}]{otsu1975threshold}
\begin{barticle}
\bauthor{\bsnm{Otsu}, \binits{N.}}, \betal:
\batitle{A threshold selection method from gray-level histograms}.
\bjtitle{Automatica}
\bvolume{11}(\bissue{285-296}),
\bfpage{23}--\blpage{27}
(\byear{1975})
\end{barticle}
\endbibitem

\bibitem[\protect\citeauthoryear{Yang et~al.}{2024}]{yang2024mambamil}
\begin{bchapter}
\bauthor{\bsnm{Yang}, \binits{S.}},
\bauthor{\bsnm{Wang}, \binits{Y.}},
\bauthor{\bsnm{Chen}, \binits{H.}}:
\bctitle{Mambamil: Enhancing long sequence modeling with sequence reordering in computational pathology}.
In: \bbtitle{International Conference on Medical Image Computing and Computer-Assisted Intervention},
pp. \bfpage{296}--\blpage{306}
(\byear{2024}).
\bcomment{Springer}
\end{bchapter}
\endbibitem

\bibitem[\protect\citeauthoryear{Klambauer et~al.}{2017}]{klambauer2017self}
\begin{botherref}
\oauthor{\bsnm{Klambauer}, \binits{G.}},
\oauthor{\bsnm{Unterthiner}, \binits{T.}},
\oauthor{\bsnm{Mayr}, \binits{A.}},
\oauthor{\bsnm{Hochreiter}, \binits{S.}}:
Self-normalizing neural networks.
Advances in neural information processing systems
\textbf{30}
(2017)
\end{botherref}
\endbibitem

\bibitem[\protect\citeauthoryear{Dalla-Torre et~al.}{2025}]{dalla2025nucleotide}
\begin{barticle}
\bauthor{\bsnm{Dalla-Torre}, \binits{H.}}, \betal:
\batitle{Nucleotide transformer: building and evaluating robust foundation models for human genomics}.
\bjtitle{Nature Methods}
\bvolume{22}(\bissue{2}),
\bfpage{287}--\blpage{297}
(\byear{2025})
\end{barticle}
\endbibitem

\bibitem[\protect\citeauthoryear{Lee et~al.}{2020}]{lee2020biobert}
\begin{barticle}
\bauthor{\bsnm{Lee}, \binits{J.}}, \betal:
\batitle{Biobert: a pre-trained biomedical language representation model for biomedical text mining}.
\bjtitle{Bioinformatics}
\bvolume{36}(\bissue{4}),
\bfpage{1234}--\blpage{1240}
(\byear{2020})
\end{barticle}
\endbibitem

\bibitem[\protect\citeauthoryear{Liu et~al.}{2023}]{liu2023m3ae}
\begin{bchapter}
\bauthor{\bsnm{Liu}, \binits{H.}}, \betal:
\bctitle{M3ae: Multimodal representation learning for brain tumor segmentation with missing modalities}.
In: \bbtitle{Proceedings of the AAAI Conference on Artificial Intelligence},
vol. \bseriesno{37},
pp. \bfpage{1657}--\blpage{1665}
(\byear{2023})
\end{bchapter}
\endbibitem

\bibitem[\protect\citeauthoryear{Vaswani et~al.}{2017}]{vaswani2017attention}
\begin{botherref}
\oauthor{\bsnm{Vaswani}, \binits{A.}}, et al.:
Attention is all you need.
Advances in neural information processing systems
\textbf{30}
(2017)
\end{botherref}
\endbibitem

\bibitem[\protect\citeauthoryear{Chen et~al.}{2020}]{chen2020simple}
\begin{bchapter}
\bauthor{\bsnm{Chen}, \binits{T.}},
\bauthor{\bsnm{Kornblith}, \binits{S.}},
\bauthor{\bsnm{Norouzi}, \binits{M.}},
\bauthor{\bsnm{Hinton}, \binits{G.}}:
\bctitle{A simple framework for contrastive learning of visual representations}.
In: \bbtitle{International Conference on Machine Learning},
pp. \bfpage{1597}--\blpage{1607}
(\byear{2020}).
\bcomment{PmLR}
\end{bchapter}
\endbibitem

\bibitem[\protect\citeauthoryear{Wilcoxon}{1992}]{wilcoxon1992individual}
\begin{bchapter}
\bauthor{\bsnm{Wilcoxon}, \binits{F.}}:
\bctitle{Individual comparisons by ranking methods}.
In: \bbtitle{Breakthroughs in Statistics: Methodology and Distribution},
pp. \bfpage{196}--\blpage{202}
(\byear{1992})
\end{bchapter}
\endbibitem

\bibitem[\protect\citeauthoryear{Lundberg and Lee}{2017}]{lundberg2017unified}
\begin{botherref}
\oauthor{\bsnm{Lundberg}, \binits{S.M.}},
\oauthor{\bsnm{Lee}, \binits{S.-I.}}:
A unified approach to interpreting model predictions.
Advances in neural information processing systems
\textbf{30}
(2017)
\end{botherref}
\endbibitem

\end{thebibliography}


\clearpage

\begin{appendices}






\section{Supplementary}
\subsection{Statistics of involved datasets}
The datasets used in this studies include the TCGA dataset for model development and internal validation, and three independent cohorts for independent validation. 
Since the TCGA dataset comprises 30 cohorts for different cancer types, we exhibited their statistics separately in Table \ref{tab:internal} for clarity.
Meanwhile, the information of independent cohorts is listed in Table \ref{tab:external}.

\subsection{Detailed experiment results}
We provided the detailed experiment results of internal validation and independent cohorts in Tables \ref{tab:res_internal} and \ref{tab:res_external}, respectively.

\subsection{Expert importance analysis}
MICE benefits from the coupled contrastive and supervised learning using pan-cancer multimodal data, which enables cross-cancer knowledge modeling. 
Figure \ref{fig:supp}a delineates these dynamics through three expert categories.
The impacts of each group of experts are evaluated by quantifying the impact of feature extracted by each group to the model’s final prediction.
The high importance of specialized experts suggests that having knowledge tailored to specific types of cancer is crucial, where unique biomarkers influence clinical decision-making for each cancer. 
Moreover, the overlapping group highlights partially shared knowledge among cancers, which has been observed considerable importance scores on multiple cohorts.
In addition, the consensual knowledge among all cancers demonstrates that consensus between different cancers are influential for predicting prognosis, proving the benefits of pre-training on pan-cancer data.
MICE dynamically balances their contributions conditioned on cancer types, enhancing the effectiveness when adapting to various clinical scenarios.

\subsection{Multimodal representation visualization}
\begin{figure*}[!ht]
\centering
    \includegraphics[width=0.99 \textwidth]{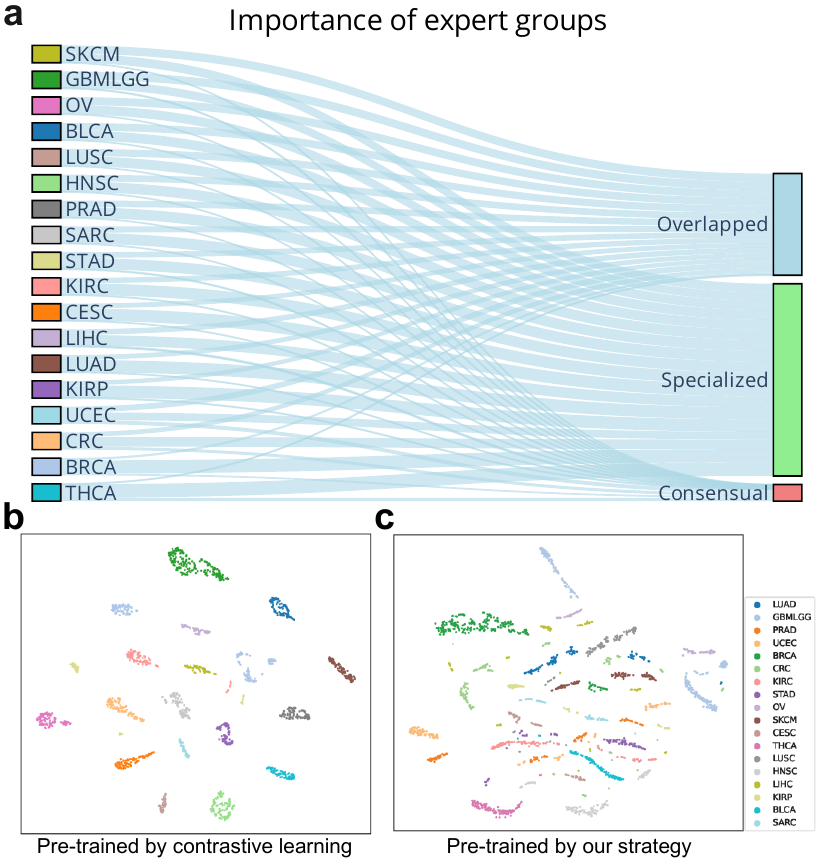}
\caption{\textbf{Model analysis on MICE.} 
\textbf{a}, The importances of different groups of experts for decision-making. 
\textbf{b-c}, Feature distributions of MICE with different pre-training strategies.
}
\label{fig:supp}
\end{figure*}

We analyzed the multimodal representation distributions extracted by MICE using the T-SNE algorithm in Figure \ref{fig:supp}b-c to intuitively validate its discrimination capacity. 
While contrastive pre-training (\ref{fig:supp}b) effectively separates representations across cancer types, it leads to overly compact intra-class distributions (e.g., tightly clustered LUAD samples), limiting discriminative power in clinical tasks focusing on discriminating patients of the same cancer type.
In contrast, MICE’s hybrid pre-training strategy conclude representations of the same cancer into multiple sub-clusters (\ref{fig:supp}c), reflecting distinct molecular or histopathological subgroups. 
This demonstrates MICE’s capability to resolve underlying biological heterogeneity among patients, a critical requirement for precision oncology applications like prognosis prediction.

\begin{table}[!t]
    \centering
    \caption{\textbf{Statistics of TCGA cohorts.} ``\# patients'' and ``\# WSIs'' indicate the numbers of patients and WSIs in each cohort, respectively. ``WSI'', ``Gene'', ``Report'', and ``Full'' indicates the numbers of patients with pathology images, molecular profiles, clinical reports, and all three modalities, respectively. $\dagger$ cohorts using progression-free survival and uncensored progression follow-ups instead of overall survival and death as others.}
    \label{tab:internal}
\begin{tabular}{l|cc|cccc|cc}
\hline
TCGA cohort       & \# patients  & \# WSIs  & WSI & Gene & Report & Full & OS     & Death             \\\hline
\multicolumn{9}{c}{Internal cohorts}                                                     \\\hline
BLCA   & 402   & 446     & 376  & 398  & 402    & 372  & 402    & 178               \\
BRCA   & 1055  & 1089    & 1023 & 1039 & 1055   & 1007 & 1055   & 151               \\
CESC   & 273   & 260     & 250  & 263  & 273    & 240  & 273    & 70                \\
CRC    & 597   & 588     & 579  & 559  & 592    & 540  & 597    & 124               \\
GBMLGG & 1002  & 1601    & 830  & 625  & 854    & 531  & 1002   & 552               \\
HNSC   & 518   & 463     & 441  & 507  & 518    & 430  & 518    & 218               \\
KIRC   & 520   & 504     & 498  & 495  & 520    & 473  & 520    & 174               \\
KIRP   & 290   & 296     & 272  & 282  & 290    & 264  & 290$\dagger$   & 58$\dagger$               \\
LIHC   & 347   & 351     & 337  & 339  & 347    & 329  & 347    & 125               \\
LUAD   & 498   & 518     & 455  & 487  & 497    & 443  & 498    & 179               \\
LUSC   & 475   & 484     & 452  & 457  & 475    & 434  & 475    & 201               \\
OV     & 326   & 101     & 101  & 291  & 290    & 66   & 326    & 191               \\
PRAD   & 497   & 448     & 402  & 493  & 478    & 398  & 497$\dagger$   & 93$\dagger$               \\
SARC   & 257   & 596     & 250  & 249  & 255    & 241  & 257    & 97                \\
SKCM   & 451   & 456     & 415  & 425  & 418    & 388  & 451    & 221               \\
STAD   & 396   & 389     & 363  & 374  & 393    & 341  & 396    & 159               \\
THCA   & 503   & 515     & 502  & 497  & 503    & 496  & 503$\dagger$   & 51$\dagger$               \\
UCEC   & 525   & 554     & 495  & 508  & 525    & 478  & 525    & 91                \\\hline
All internal & 8932  & 9659    & 8041 & 8288 & 8685   & 7471 & 8932   & 2933              \\\hline
\multicolumn{9}{c}{Augmented cohorts}                                                    \\\hline
ACC    & 80    & 227     & 56   & 78   & 80     & 54   & 80     & 28                \\
CHOL   & 36    & 36      & 36   & 33   & 36     & 33   & 36     & 17                \\
DLBC   & 45    & 41      & 41   & 45   & 45     & 41   & 45     & 9                 \\
ESCA   & 177   & 153     & 151  & 175  & 177    & 149  & 177    & 74                \\
KICH   & 104   & 116     & 104  & 65   & 104    & 65   & 104    & 10                \\
MESO   & 84    & 84      & 72   & 84   & 84     & 72   & 84     & 72                \\
PAAD   & 179   & 203     & 177  & 171  & 170    & 162  & 179    & 99                \\
PCPG   & 172   & 189     & 169  & 171  & 171    & 167  & 172    & 6                 \\
TGCT   & 130   & 207     & 129  & 129  & 125    & 124  & 130    & 3                 \\
THYM   & 122   & 179     & 119  & 117  & 120    & 114  & 122    & 9                 \\
UCS    & 55    & 85      & 55   & 55   & 55     & 55   & 55     & 34                \\
UVM    & 75    & 75      & 75   & 75   & 75     & 75   & 75     & 23                \\\hline
All augmented & 1259  & 1595    & 1184 & 1198 & 1242   & 1111 & 1259   & 384               \\\hline
All TCGA       & 10191 & 11254   & 9225 & 9486 & 9927   & 8582 & 10191  & 3317              \\\hline
\end{tabular}
\end{table}

\begin{table}[!ht]
    \centering
    \caption{\textbf{Statistics of independent cohorts.} ``\# patients'' and ``\# WSIs'' indicate the numbers of patients and WSIs in each cohort, respectively. OS: overall survival; DFS: disease-free survival; DSS: disease-specific survival; PFS: progression-free survival; DMFS: distant metastasis-free survival; LRFS: local recurrence-free survival.}
    \label{tab:external}
\setlength{\tabcolsep}{17pt}
\begin{tabular}{c|c|c|c}
\hline
Cohort             & HANCOCK           & YNGC              & NFCRC                               \\\hline
Availability       & Public            & Private           & Private                          \\\hline
Cancer type        & Head \& neck      & Gastric           & Colorectal                       \\\hline
\# patients        & 726               & 580               & 302                                 \\\hline
\# WSIs            & 726               & 580               & 316                                 \\\hline
\multirow{2}{*}{Modality}      & Pathology image,     & Pathology image,     & Pathology image,            \\
                   & surgery report      & pathology report      & pathology report               \\\hline
\multirow{5}{*}{\makecell{Follow-ups\\(Uncensored/all)}}     & OS (202/726)     &                   & \multirow{5}{*}{\makecell{OS (43/302) \\DSS (42/301)}}               \\
   & DFS (169/726)    & OS (206/580)     &                 \\
                   & DSS(107/717)     & LRFS (47/580)      &                                           \\
                   & PFS (166/726)    & DMFS (136/579)      &                                           \\
                   & DMFS (103/725)      &                   &                                           \\\hline
\end{tabular}
\end{table}

\begin{table}[htbp]
\centering
\caption{\textbf{Performance of prognosis prediction on internal TCGA cohorts.} We exhibit the mean C-index and standard deviation with 95\% confidence interval in parentheses. Best scores are in \textbf{bold}.}
\label{tab:res_internal}
\setlength{\tabcolsep}{3pt}
\tiny
\begin{tabular}{l|ccc|ccccc}
\hline
TCGA        & \multicolumn{3}{c|}{Unimodal models} & \multicolumn{5}{c}{Multimodal models} \\\cline{2-9}
cohort      & Report & Pathology & Genomics & Early & Late  & MulHead \cite{shao2024multi} & MoE \cite{jiang2024m4oe}   & MICE \\\hline
BLCA & 0.536±0.052 & 0.626±0.036 & 0.630±0.032 & 0.637±0.079 & 0.621±0.056 & 0.641±0.056 & 0.653±0.050 & \textbf{0.673±0.052} \\
& (0.445, 0.644) & (0.533, 0.716) & (0.546, 0.703) & (0.540, 0.746) & (0.532, 0.707) & (0.561, 0.741) & (0.562, 0.751) & (0.575, 0.764) \\\hline
BRCA & 0.589±0.077 & 0.644±0.024 & 0.691±0.025 & 0.659±0.094 & 0.642±0.094 & 0.654±0.067 & 0.663±0.087 & \textbf{0.733±0.075} \\
& (0.485, 0.702) & (0.524, 0.737) & (0.571, 0.793) & (0.528, 0.768) & (0.514, 0.731) & (0.518, 0.770) & (0.554, 0.742) & (0.612, 0.830) \\ \hline
CESC & 0.546±0.046 & 0.620±0.080 & 0.648±0.048 & 0.637±0.046 & 0.649±0.042 & 0.681±0.064 & 0.663±0.051 & \textbf{0.724±0.081} \\
& (0.381, 0.655) & (0.451, 0.745) & (0.465, 0.764) & (0.444, 0.779) & (0.476, 0.762) & (0.495, 0.789) & (0.487, 0.807) & (0.546, 0.850) \\ \hline
CRC & 0.576±0.130 & 0.657±0.042 & 0.622±0.038 & 0.638±0.069 & 0.613±0.032 & 0.674±0.025 & 0.630±0.046 & \textbf{0.693±0.085} \\
 & (0.454, 0.669) & (0.527, 0.768) & (0.493, 0.746) & (0.513, 0.755) & (0.456, 0.727) & (0.540, 0.781) & (0.483, 0.754) & (0.553, 0.801) \\ \hline
GBMLGG & 0.661±0.036 & 0.724±0.023 & 0.709±0.044 & \textbf{0.774±0.028} & 0.750±0.017 & 0.745±0.027 & 0.765±0.017 & 0.771±0.019 \\
 & (0.612, 0.724) & (0.691, 0.766) & (0.651, 0.751) & (0.728, 0.834) & (0.699, 0.783) & (0.704, 0.794) & (0.732, 0.796) & (0.723, 0.814) \\ \hline
HNSC & 0.571±0.064 & 0.646±0.054 & 0.633±0.047 & 0.614±0.016 & 0.628±0.035 & 0.647±0.046 & \textbf{0.692±0.040} & 0.690±0.050 \\
 & (0.500, 0.665) & (0.547, 0.726) & (0.535, 0.709) & (0.534, 0.685) & (0.539, 0.725) & (0.550, 0.744) & (0.595, 0.755) & (0.599, 0.770) \\ \hline
KIRC & 0.683±0.045 & 0.718±0.049 & 0.662±0.022 & 0.694±0.021 & 0.730±0.051 & \textbf{0.760±0.064} & 0.732±0.041 & \textbf{0.760±0.061} \\
 & (0.598, 0.753) & (0.616, 0.778) & (0.549, 0.731) & (0.596, 0.746) & (0.643, 0.804) & (0.664, 0.816) & (0.626, 0.788) & (0.662, 0.829) \\ \hline
KIRP & 0.637±0.095 & 0.674±0.074 & 0.699±0.056 & 0.706±0.098 & 0.769±0.079 & 0.783±0.075 & 0.790±0.082 & \textbf{0.830±0.066} \\
 & (0.445, 0.732) & (0.501, 0.803) & (0.507, 0.795) & (0.539, 0.820) & (0.589, 0.879) & (0.591, 0.882) & (0.595, 0.920) & (0.652, 0.941) \\ \hline
LIHC & 0.648±0.068 & 0.645±0.016 & 0.659±0.030 & 0.713±0.038 & 0.742±0.051 & 0.724±0.045 & 0.751±0.044 & \textbf{0.761±0.045} \\
 & (0.525, 0.744) & (0.538, 0.757) & (0.532, 0.756) & (0.613, 0.810) & (0.625, 0.843) & (0.598, 0.810) & (0.630, 0.857) & (0.644, 0.861) \\ \hline
LUAD & 0.597±0.034 & 0.651±0.013 & 0.627±0.058 & 0.630±0.051 & 0.627±0.015 & 0.668±0.016 & 0.654±0.053 & \textbf{0.717±0.019} \\
 & (0.452, 0.670) & (0.495, 0.709) & (0.471, 0.717) & (0.506, 0.711) & (0.498, 0.696) & (0.517, 0.755) & (0.525, 0.721) & (0.577, 0.790) \\ \hline
LUSC & 0.573±0.054 & 0.552±0.039 & 0.569±0.038 & 0.583±0.019 & 0.536±0.026 & 0.569±0.044 & 0.576±0.010 & \textbf{0.621±0.035} \\
 & (0.478, 0.680) & (0.446, 0.634) & (0.486, 0.652) & (0.476, 0.695) & (0.442, 0.624) & (0.488, 0.675) & (0.494, 0.658) & (0.521, 0.720) \\ \hline
OV & 0.623±0.035 & 0.592±0.048 & 0.592±0.048 & \textbf{0.650±0.042} & 0.615±0.018 & 0.598±0.030 & 0.632±0.023 & 0.643±0.023 \\
& (0.489, 0.724) & (0.457, 0.691) & (0.469, 0.697) & (0.529, 0.736) & (0.504, 0.724) & (0.485, 0.679) & (0.513, 0.732) & (0.523, 0.739) \\ \hline
PRAD & 0.616±0.069 & 0.654±0.065 & 0.648±0.067 & 0.606±0.085 & \textbf{0.718±0.044} & 0.672±0.038 & 0.683±0.022 & 0.717±0.059 \\
& (0.449, 0.705) & (0.521, 0.744) & (0.512, 0.769) & (0.466, 0.712) & (0.580, 0.830) & (0.520, 0.791) & (0.527, 0.788) & (0.566, 0.822) \\ \hline
SARC & 0.577±0.063 & 0.634±0.063 & 0.659±0.031 & 0.695±0.021 & 0.625±0.044 & 0.623±0.053 & 0.690±0.020 & \textbf{0.698±0.037} \\
 & (0.398, 0.704) & (0.488, 0.751) & (0.501, 0.781) & (0.529, 0.821) & (0.462, 0.742) & (0.473, 0.753) & (0.528, 0.828) & (0.535, 0.820) \\ \hline
SKCM & 0.552±0.034 & 0.595±0.036 & 0.619±0.041 & 0.621±0.039 & 0.606±0.024 & 0.620±0.043 & \textbf{0.634±0.010} & 0.629±0.033 \\
& (0.478, 0.660) & (0.501, 0.691) & (0.531, 0.731) & (0.512, 0.711) & (0.514, 0.706) & (0.512, 0.726) & (0.561, 0.734) & (0.536, 0.723) \\ \hline
STAD & 0.601±0.061 & 0.566±0.028 & 0.567±0.027 & 0.586±0.025 & 0.568±0.040 & 0.595±0.024 & 0.585±0.045 & \textbf{0.653±0.028} \\
& (0.471, 0.692) & (0.431, 0.633) & (0.428, 0.649) & (0.438, 0.651) & (0.437, 0.654) & (0.478, 0.698) & (0.460, 0.686) & (0.517, 0.737) \\ \hline
THCA & 0.618±0.050 & 0.610±0.108 & 0.628±0.037 & 0.572±0.038 & 0.642±0.103 & 0.690±0.085 & 0.637±0.063 & \textbf{0.720±0.075} \\
& (0.428, 0.753) & (0.426, 0.745) & (0.467, 0.773) & (0.390, 0.733) & (0.467, 0.784) & (0.528, 0.844) & (0.452, 0.770) & (0.541, 0.868) \\ \hline
UCEC & 0.563±0.081 & 0.668±0.036 & 0.657±0.026 & 0.611±0.073 & 0.717±0.068 & \textbf{0.745±0.065} & 0.671±0.066 & 0.742±0.048 \\
& (0.381, 0.652) & (0.516, 0.745) & (0.485, 0.762) & (0.437, 0.702) & (0.538, 0.827) & (0.585, 0.837) & (0.516, 0.769) & (0.573, 0.839) \\ \hline
Mean & 0.598 & 0.638 & 0.640 & 0.646 & 0.655 & 0.672 & 0.672 & \textbf{0.710} \\
\hline
\end{tabular}
\end{table}

\begin{table}[htbp]
\centering
\caption{\textbf{Performance of prognosis prediction on independent cohorts.} We exhibit the mean C-index and standard deviation with 95\% confidence interval in parentheses. OS: overall survival; DFS: disease-free survival; DSS: disease-specific survival; PFS: progression-free survival; DMFS: distant metastasis-free survival; LRFS: local recurrence-free survival. Best scores are in \textbf{bold}.}
\setlength\tabcolsep{1pt}
\label{tab:res_external}
\tiny
\begin{tabular}{c|cc|ccccccc}
\hline
Cohort & \multicolumn{2}{c|}{Unimodal models} & \multicolumn{7}{c}{Multimodal models} \\\cline{2-10}
(Task)                & Report & Pathology  & Early & Late & MulHead \cite{shao2024multi} & MoE \cite{jiang2024m4oe}   & MCAT \cite{mcat}  & Survpath \cite{jaume2024modeling} & MICE \\\hline
NFCRC & 0.664±0.024 & 0.683±0.016 & 0.689±0.020 & 0.680±0.034 & 0.685±0.037 & 0.701±0.051 & 0.688±0.058 & 0.693±0.035 & \textbf{0.776±0.058} \\
(OS) & (0.507, 0.801) & (0.536, 0.828) & (0.530, 0.845) & (0.512, 0.825) & (0.505, 0.819) & (0.542, 0.832) & (0.522, 0.843) & (0.548, 0.831) & (0.615, 0.917) \\ \hline
NFCRC & 0.633±0.074 & 0.681±0.034 & 0.731±0.074 & 0.679±0.081 & 0.703±0.083 & 0.729±0.019 & 0.717±0.038 & 0.727±0.059 & \textbf{0.789±0.039} \\
(DSS) & (0.510, 0.762) & (0.559, 0.832) & (0.586, 0.861) & (0.530, 0.824) & (0.550, 0.851) & (0.574, 0.860) & (0.569, 0.846) & (0.571, 0.852) & (0.647, 0.923) \\ \hline
YNGC & 0.579±0.030 & 0.598±0.087 & 0.613±0.074 & 0.591±0.038 & 0.625±0.074 & 0.610±0.079 & 0.614±0.048 & 0.580±0.062 & \textbf{0.665±0.034} \\
(OS) & (0.481, 0.641) & (0.524, 0.684) & (0.527, 0.688) & (0.503, 0.680) & (0.546, 0.701) & (0.511, 0.685) & (0.537, 0.685) & (0.502, 0.668) & (0.583, 0.743) \\ \hline
YNGC & 0.549±0.084 & 0.577±0.066 & 0.565±0.035 & 0.561±0.085 & 0.577±0.069 & 0.575±0.053 & 0.535±0.040 & 0.563±0.051 & \textbf{0.674±0.050} \\
(LRFS) & (0.361, 0.712) & (0.395, 0.742) & (0.379, 0.698) & (0.402, 0.711) & (0.410, 0.732) & (0.406, 0.721) & (0.352, 0.679) & (0.382, 0.721) & (0.503, 0.820) \\ \hline
YNGC & 0.568±0.045 & 0.593±0.051 & 0.576±0.049 & 0.569±0.042 & 0.597±0.057 & 0.581±0.058 & 0.549±0.068 & 0.598±0.042 & \textbf{0.715±0.041} \\
(DMFS) & (0.468, 0.638) & (0.513, 0.699) & (0.462, 0.670) & (0.466, 0.659) & (0.520, 0.668) & (0.469, 0.668) & (0.444, 0.643) & (0.508, 0.675) & (0.620, 0.801) \\ \hline
HANC & 0.653±0.049 & 0.663±0.050 & 0.675±0.052 & 0.679±0.045 & 0.678±0.072 & 0.664±0.075 & 0.675±0.052 & 0.687±0.048 & \textbf{0.727±0.044} \\
(DMFS) & (0.527, 0.748) & (0.544, 0.775) & (0.574, 0.775) & (0.585, 0.774) & (0.561, 0.797) & (0.546, 0.777) & (0.559, 0.769) & (0.561, 0.781) & (0.619, 0.837) \\ \hline
HANC & 0.661±0.055 & 0.662±0.039 & 0.657±0.049 & 0.678±0.062 & 0.666±0.052 & 0.651±0.042 & 0.659±0.032 & 0.670±0.057 & \textbf{0.700±0.057} \\
(DSS) & (0.567, 0.763) & (0.561, 0.784) & (0.538, 0.768) & (0.574, 0.803) & (0.536, 0.764) & (0.520, 0.742) & (0.542, 0.779) & (0.574, 0.773) & (0.587, 0.805) \\ \hline
HANC & 0.619±0.036 & 0.622±0.035 & 0.630±0.042 & 0.643±0.036 & 0.640±0.072 & 0.632±0.063 & 0.631±0.035 & 0.635±0.027 & \textbf{0.663±0.030} \\
(OS) & (0.545, 0.716) & (0.521, 0.716) & (0.541, 0.734) & (0.569, 0.726) & (0.557, 0.713) & (0.548, 0.723) & (0.550, 0.705) & (0.525, 0.742) & (0.570, 0.755) \\ \hline
HANC & 0.599±0.051 & 0.603±0.055 & 0.601±0.053 & 0.615±0.065 & 0.620±0.073 & 0.590±0.086 & 0.609±0.046 & 0.617±0.065 & \textbf{0.636±0.023} \\
(DFS) & (0.483, 0.681) & (0.486, 0.676) & (0.492, 0.693) & (0.478, 0.712) & (0.502, 0.724) & (0.488, 0.679) & (0.504, 0.703) & (0.488, 0.709) & (0.519, 0.722) \\ \hline
HANC & 0.587±0.034 & 0.615±0.025 & 0.620±0.059 & 0.626±0.040 & 0.621±0.073 & 0.611±0.082 & 0.610±0.053 & 0.624±0.028 & \textbf{0.648±0.025} \\
(PFS) & (0.492, 0.711) & (0.525, 0.723) & (0.532, 0.732) & (0.522, 0.739) & (0.545, 0.728) & (0.540, 0.708) & (0.522, 0.726) & (0.531, 0.736) & (0.557, 0.758) \\ \hline
Mean & 0.611 & 0.630 & 0.636 & 0.632 & 0.641 & 0.634 & 0.629 & 0.639 & \textbf{0.699} \\
\hline
\end{tabular}
\end{table}

\end{appendices}

\end{document}